\documentclass[12pt]{article}

\usepackage{newtxtext,newtxmath}
\usepackage{graphicx}

\usepackage{booktabs}

\usepackage[letterpaper,margin=1in]{geometry}
\usepackage{siunitx}
\usepackage{scicite}
\usepackage{hyperref}
\usepackage{cleveref}
\usepackage{wrapfig}
\usepackage{caption}
\usepackage[CaptionAfterwards]{fltpage}

\renewenvironment{abstract}
	{\quotation}
	{\endquotation}

\date{}

\makeatletter
\renewcommand{\fnum@figure}{\textbf{Figure \thefigure}}
\renewcommand{\fnum@table}{\textbf{Table \thetable}}
\makeatother

\usepackage{scicite}

\usepackage{url}
\usepackage{xcolor}

\newcommand{\vect}[1]{\mathbf{#1}}   % vectors
\newcommand{\mat}[1]{\mathbf{#1}}    % matrices
\newcommand{\gvect}[1]{\boldsymbol{#1}} % Greek vectors

\usepackage{cleveref}
\usepackage[normalem]{ulem}

\def\scititle{
	Embodied Passive Aeroacoustic Perception Enables Relative Sensing and Pursuit Between Aerial Robots

}
\title{\bfseries \boldmath \scititle}

\author{
    Yanbaihui Liu$^{1}$,
    Ravi Prakash$^{1}$,
    Li-Yu Lo$^{1}$,
    Nils Roede$^{1}$,
    Boyuan Chen$^{1,2,3\ast}$\and
    \small$^{1}$Department of Mechanical Engineering and Materials Science, Duke University, Durham, 27708, USA.\and
    \small$^{2}$Department of Electrical and Computer Engineering, Duke University, Durham, 27708, USA.\and
    \small$^{3}$Department of Computer Science, Duke University, Durham, 27708, USA.\and
    \small$^\ast$To whom correspondence should be addressed; E-mail: boyuan.chen@duke.edu.
}

\newcommand{\projectwebsite}{%
  \begin{center}
    \vspace{-0.5cm}
    \href{http://generalroboticslab.com/SonicFly}{\textcolor{orange}{http://generalroboticslab.com/SonicFly}}
  \end{center}
}

\begin{document} 

% Insert the title and author list
\maketitle

\projectwebsite

% Abstract, in bold
% There are strict length limits, and not all formats have abstracts.
% Consult the journal instructions to authors for details.
% Do not cite any references in the abstract.
\begin{abstract} \bfseries \boldmath
Aerial robots continuously generate structured aeroacoustic fields during flight, yet these signals have been largely underexplored as a source of onboard relative perception, particularly under the strong ego-acoustic interference generated during simultaneous flight in various outdoor conditions. We introduce \textit{embodied passive aeroacoustic perception}, a sensing paradigm in which an aerial robot infers actionable relative-state information from the naturally generated sound of flight while operating within its own evolving aeroacoustic field. We present SonicFly, a passive aeroacoustic perception framework that enables one unmanned aerial vehicle (UAV) to estimate and follow another using only the leader's intrinsic flight sound, without active acoustic signaling, inter-robot communication, GPS sharing, or external sensing infrastructure. The system uses a lightweight four-microphone array, rotorcraft-informed acoustic representations, a neural bearing-range estimator, and confidence-gated filtering for closed-loop flight. Through acoustic characterization, onboard localization, and outdoor pursuit experiments, we show that multirotor aeroacoustic signals contain sufficient information to support relative perception despite strong ego-acoustic interference, environmental variability, and continuously changing flight geometry. During acoustic-only pursuit, SonicFly achieved a mean distance-maintenance error of 1.34 m across diverse outdoor trajectories and operating conditions. Analysis of the acoustic channel further reveals design principles governing embodied passive aeroacoustic perception, including the roles of harmonic structure, spectral separability, and spatial acoustic cues in determining observability. Our results establish the feasibility of embodied passive aeroacoustic perception for aerial robots and suggest that naturally generated behavioral signals can serve as useful information for robotic perception and coordination.
\end{abstract}

\section*{Summary} 
Aerial robots can estimate and pursue one another using only the naturally generated sound of flight.

% The first paragraph of any Science paper does NOT have a heading
% Nor is it indented
\section*{Introduction}

Animals continuously generate signals as a consequence or byproduct of their movements. Recent biological studies have discovered that animals can leverage such signals for coordination and communication. Aquatic insects communicate through surface ripples \cite{wilcox1995ripple}, flying animals generate structured aerodynamic sounds \cite{clark2020humming}, fish sense hydrodynamic disturbances produced by nearby swimmers \cite{ko2023role, scott2023lateral}, and social fishes exchange information through subtle body-induced mechanosensory cues \cite{butler2016mechanosensory}. Across these systems, coordination does not require explicit messages, shared communication protocols, or direct visual contact. Instead, animals can exploit information embedded in the naturally generated consequences of behavior. These passive signals often provide robust and energy-efficient sources of information, enabling coordination in darkness, cluttered environments, and other sensing-degraded conditions.

Despite these biological precedents, modern robotic systems rarely exploit such naturally generated signals from behaviors and instead often rely heavily on active sensing, communication, or line-of-sight-dependent perception. Satellite-based methods provide globally referenced positions for multi-robot coordination, acting as a third-party reference in open-sky conditions \cite{schaefer2021accuracy,um2020configuring} but become unreliable in Global Positioning System (GPS)-denied settings such as urban canyons, dense canopies, or tunnels, and are inherently susceptible to intentional jamming and spoofing \cite{xie2014measuring, ioannides2016known, psiaki2016gnss}. Radio-frequency ranging methods, including ultra-wideband (UWB), offer an alternative high-relative-estimation solution but require active packet exchange between cooperative transceivers \cite{kempke2015polypoint, ledergerber2015robot, tiemann2017scalable, shule2020uwb}, introducing strict hardware, synchronization, and communication dependencies. Meanwhile, onboard exteroceptive perception relies heavily on fiducial markers \cite{xiao2017visual, faessler2014monocular, lo2024experimental}, learning-based visual detectors \cite{redmon2016you, ren2015faster, carion2020end, bonatti2019towards, lo2021dynamic, xu2022omni}, and LiDAR-based target localization and tracking \cite{zhou2018voxelnet,weng20203d,vrba2024onboard,zhu2024swarm} to support relative localization and cooperative operation. While these modalities have enabled significant advances in aerial autonomy, they share a critical vulnerability in their common reliance on external infrastructure, active continuous signal exchange, or pristine sensing conditions, qualities that can be hard to assure in real-world scenarios. This suggests a complementary direction in which robots sense from signals that are already present during operation, rather than from signals or references introduced for the sole objective of sensing.

\begin{figure}
\centering
\includegraphics[width=1\textwidth]{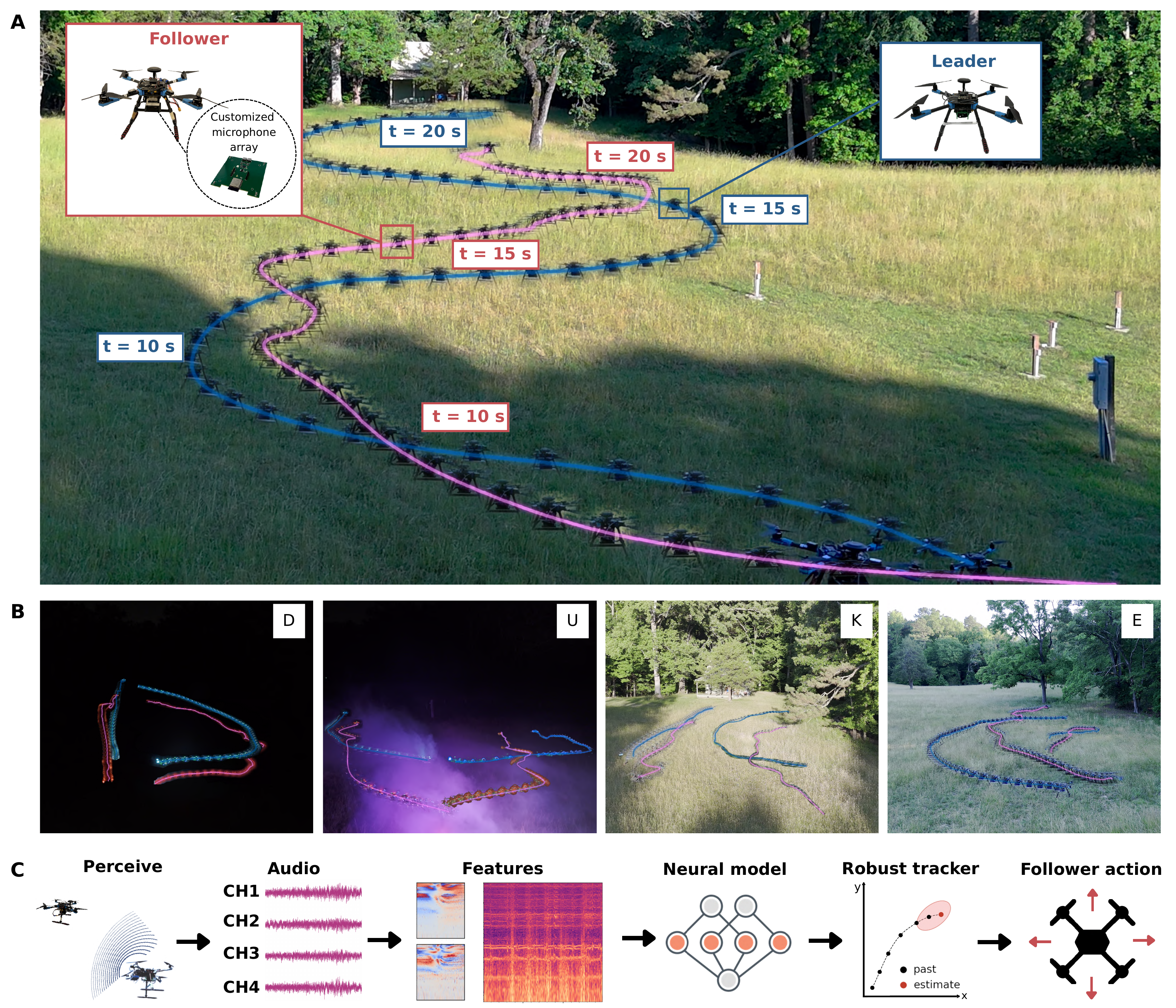}
\caption{\textbf{Passive aeroacoustic perception enables onboard drone tracking and pursuit.} 
(\textbf{A}) SonicFly uses an onboard microphone array to passively sense the aeroacoustic signature of a leader drone and estimate its relative motion during pursuit. 
(\textbf{B}) Acoustic-only pursuit is evaluated across diverse outdoor conditions, including darkness, fog, strong sunlight, and cloudy scenes. 
(\textbf{C}) SonicFly converts four-channel audio into spectral and spatial acoustic features, including spectrogram, IPD, and ILD. These features are processed by SonicNet and stabilized by a robust tracker to estimate the leader state and generate the actions for the follower drone.}
\label{fig:sonicfly_overview}
\end{figure}

For multirotor unmanned aerial vehicles (UAVs), a particularly ubiquitous but underexplored signal is the sound of flight itself. UAVs continuously produce complex but highly structured aeroacoustic fields through rotor motion, aerodynamic loading, turbulent wake interactions, and airframe vibrations \cite{salvati2019acoustic, djurek2020analysis}. These flight-induced acoustic signatures contain rich harmonic structure determined by rotor geometry, rotational speed, and flight dynamics. Earlier work has demonstrated that drone-generated sounds can support acoustic localization and relative bearing estimation under sensing conditions that avoid self-engine interference in static and indoor environments \cite{basiri2016board}, whereas most aerial-robot acoustic studies have largely treated rotor sound as ego-noise to be mitigated or as an acoustic emission to be reduced \cite{prakhar2025bioinspired, gupta2026droneaudioset}.
Other existing robotic acoustic systems predominantly rely on active sensing, including sonar, ultrasound, acoustic probing, and emitted acoustic signals \cite{basiri2013audio,basiri2014audio,libby2021multiclass,liu2024sonicsense,liu2025wildfusion,calkins2021distance,schroeder2024superbat,velmurugan2026milliwatt}. Additional efforts that leverage intrinsic drone sounds focus strictly on acoustic recognition for detection \cite{chatterjee2025audron} or external microphone infrastructure~\cite{sun2022aim,chen2022boombox}, both in controlled indoor environments. However, a fundamentally different sensing paradigm remains largely unexplored: can a flying aerial robot continuously estimate and coordinate with another using only naturally generated sound of flight while simultaneously operating inside its own dynamically evolving ego-acoustic field (fig.~\ref{fig:sonicfly_overview})?

Realizing such a capability is challenging. Unlike previous onboard acoustic localization settings, the observing UAV in our formulation is itself continuously flying and therefore immersed within a strong, dynamically changing ego-acoustic field. The observer itself is therefore a powerful acoustic source. In a two-drone tracking and pursuit setting, we refer to the observing drone as the ``follower'' and the target drone as the ``leader''. Unlike conventional drone-sound detection, where microphones are often stationary, bulky, and acoustically isolated from the target, an onboard listener must operate within its own ego-acoustic field. During flight, the follower continuously generates strong acoustic energy from its own propellers, motors, and aerodynamic wake, often occupying similar frequency bands that contain the information of the leader. The leader's acoustic signal further attenuates with distance and mixes with the follower's self-noise before reaching the onboard microphones. Moreover, wind, humidity, temperature, ambient sound, and continuously changing leader-follower geometry all influence acoustic propagation. As quantified in Figure~\ref{fig:drones_acoustic_signature}A, although leader-related harmonic structure remains noticeable across listening conditions, embodied flight and wind substantially reduce peak signal-to-noise ratio (SNR), making leader signals sparse compared with the follower's broadband ego acoustics. Flying motions also make the acoustic field inherently nonstationary. Changes in attitude, thrust, downwash, wind direction, and leader-follower geometry all alter both spectral content and the interaural phase differences (IPD) and level differences (ILD) that carry spatial information.

Therefore, flying with a leader drone purely based on passive acoustics turns passive listening into a much more challenging embodied perception problem. The objective is not simply to detect drone sound, but to continuously recover actionable relative leader-follower state information from an aeroacoustic field dominated by the follower itself. The task also requires tightly coupled perception and action considerations to maintain a stable team structure under rapidly evolving conditions. We define this problem as \textit{embodied passive aeroacoustic perception}: estimating the relative state of a leader UAV from the aeroacoustics naturally generated during flight and using these estimates to guide subsequent motion of a follower UAV, in order to maintain a relatively stable leader-follower teaming structure. Unlike external drone detection, which typically assumes a stationary or acoustically isolated listener, the follower UAV is itself an active participant in the sensing process. It must infer state from residual spectral, phase, and level signatures embedded within its own ego-aeroacoustic field while simultaneously using these estimates to determine its next action.

This formulation differs fundamentally from conventional passive acoustic localization and active acoustic sensing. In active acoustic systems~\cite{velmurugan2026milliwatt}, a robot emits probing signals and estimates spatial structure from acoustic returns. Conventional passive acoustic localization neither operates within a flying robot’s ego-aeroacoustic field \cite{basiri2016board,sun2022aim} nor handles outdoor environments without relying on spectrally distinct acoustic signals \cite{basiri2013audio,basiri2014audio}. In embodied passive aeroacoustic perception, the follower instead remains continuously airborne while relying entirely on naturally generated flight sound to estimate and follow the leader. Every control action alters the subsequent source-receiver geometry, acoustic interference, and observability of the leader. The act of flight therefore creates the signal, perturbs the signal, and determines what will be sensed next. Perception and action become inseparable components of a closed-loop sensing process.

We introduce \textbf{\textit{SonicFly}}, a passive aeroacoustic perception framework that enables a follower UAV to estimate and follow another using only the leader's intrinsic flight sound during simultaneous flight of both vehicles (Fig.~\ref{fig:sonicfly_overview}). Our key insight is that the aeroacoustic field generated by multirotor UAV flight contains structured spectral and spatial information that can encode actionable relative state. SonicFly requires no GPS sharing, inter-robot communication, fiducial markers, active acoustic probing, external sensing infrastructure, or controlled indoor environments. Instead, it leverages naturally generated aeroacoustic signatures as a passive information channel for relative perception. SonicFly combines a lightweight and customized onboard four-microphone array, rotorcraft-informed acoustic representations, a deep neural bearing-range estimator, and confidence-gated temporal filtering to recover relative leader-follower state information under strong ego-acoustic interference (fig.~\ref{fig:sonicfly_overview}C).

We demonstrate SonicFly in an aerial pursuit task in which a follower UAV estimates and follows a leader UAV using passive aeroacoustic observations alone. Importantly, our goal is not to establish a universal replacement for cameras, LiDAR, GPS, or communication-based localization. Rather, we seek to demonstrate the feasibility and potential of embodied passive aeroacoustic perception as a potential sensing paradigm for aerial robots. Through acoustic characterization, onboard localization experiments, and outdoor pursuit under diverse settings (ego noise, wind, temperature, and platform variations) and environmental conditions (darkness, fog, strong sunlight, and cloudy scenes (fig.~\ref{fig:sonicfly_overview}B)), we investigate how naturally generated flight sound can be transformed from a nuisance signal into an actionable source of relative-state information. Our system achieves a mean distance error of \SI{1.34}{\meter} (target distance: \SI{3.5}{\meter}; drone diagonal length: \SI{0.5}{\meter}) during aeroacoustic-only pursuit using onboard passive observations alone. Beyond the specific system presented here, our work provides a demonstration of embodied passive aeroacoustic perception in aerial robotic systems together with an analysis of its design principles, opportunities, and limitations. More broadly, our results suggest that naturally generated signals during motion, often treated as nuisance effects or self-noise, can instead serve as useful information for robotic perception and coordination.

\section*{Results}

\subsection*{Conditions that enable embodied passive aeroacoustic perception}

\begin{FPfigure}
    \centering
    \includegraphics[width=\linewidth]{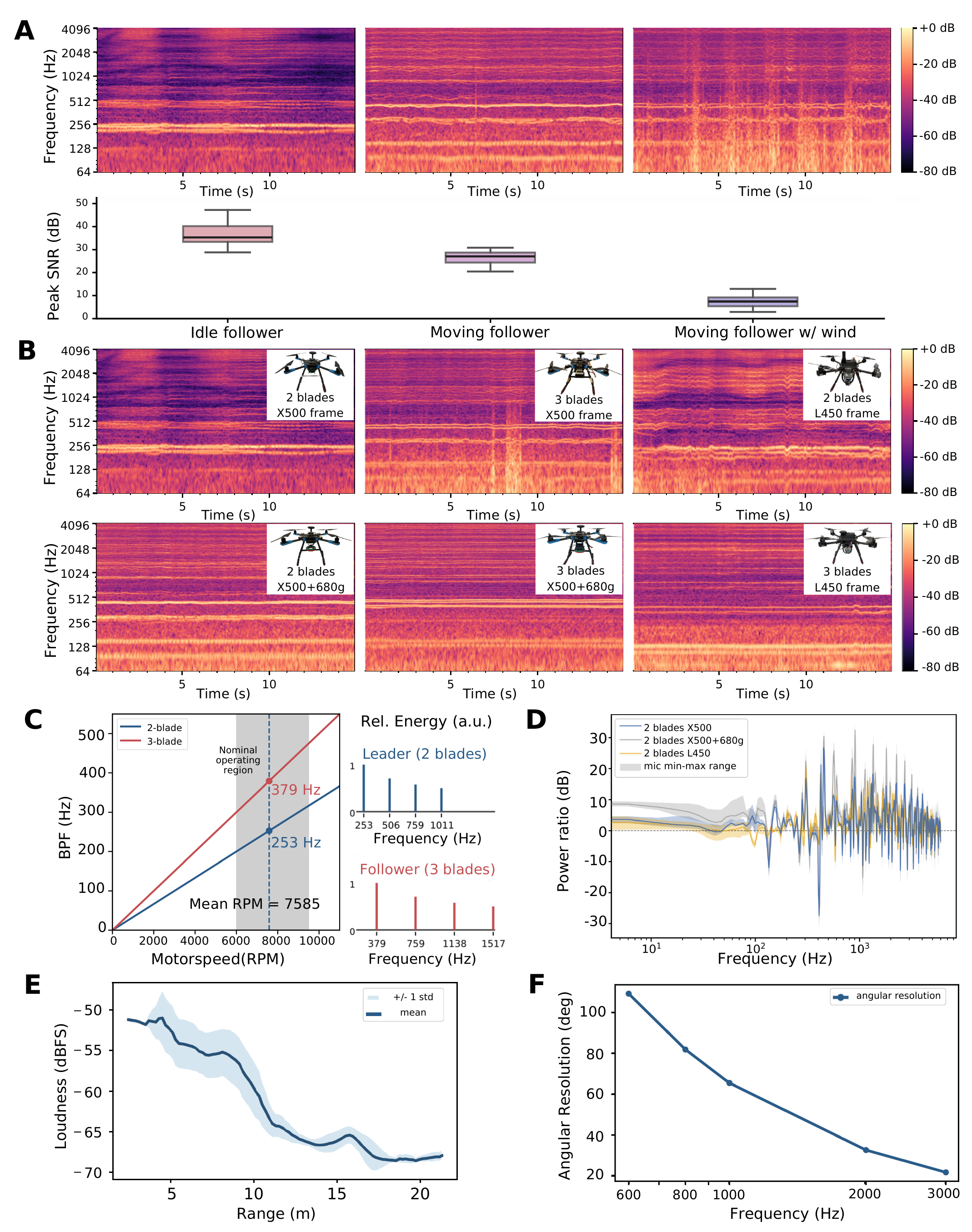}
    \caption{\textbf{Rotorcraft configuration shapes multirotor acoustic signatures.}
    (\textbf{A}) Representative spectrograms and peak SNR distributions across idle follower, moving follower, and moving follower with wind conditions demonstrate the degradation of acoustic observability during flight.
    (\textbf{B}) Representative spectrograms for six multirotor platform variants show platform-dependent tonal bands and harmonic structure.
    (\textbf{C}) Theoretical blade-passing frequency versus motor speed and blade number, with the nominal operating region measured during flight, and corresponding harmonic frequencies for the two-blade leader and three-blade follower.
    (\textbf{D}) Experimentally measured power spectral density ratio relative to the hovering follower drone. Lines indicate channel means, and shaded bands indicate variation across microphones.
    (\textbf{E}) Experimental loudness decreases with range.
    (\textbf{F}) Theoretical angular resolution of the microphone array improves at higher acoustic frequencies.}
    \label{fig:drones_acoustic_signature}
\end{FPfigure}

Embodied passive aeroacoustic perception is fundamentally different from conventional drone-sound detection. In the proposed formulation, the sensing platform is itself a strong acoustic source whose motion continuously alters the acoustic field being measured. Before evaluating localization and pursuit, we therefore first investigate whether multirotor flight acoustics contain sufficient structure to support relative perception and identify the physical conditions under which such a sensing paradigm becomes feasible.

\subsubsection*{Flight acoustics contain structured and measurable information}

Multirotor flight produces highly structured acoustic signatures rather than featureless broadband noise. The dominant components arise from periodic blade passage, with additional harmonics and broadband energy generated by aerodynamic loading, motor vibration, and rotor-airframe interactions. For a rotor with $N_b$ blades rotating at $\Omega$ revolutions per minute (RPM), the blade-passing frequency (BPF) is
\begin{equation}
f_{\mathrm{BPF}}=\frac{N_b\Omega}{60},
\end{equation}
with harmonics at integer multiples of the BPF. This relationship links the mechanical configuration of a rotorcraft to the frequency bands that a passive listener should expect. In practice, these bands shift with thrust and attitude regulation, but the spectrograms retain persistent harmonic structure associated with blade count, motor speed, loading, and rotor-frame-propeller interactions.

Figure~\ref{fig:drones_acoustic_signature}B shows representative spectrograms collected from six variations of the multirotor platforms. Distinct tonal bands and harmonic stacks were consistently observed despite differences in airframe and propulsion configurations. The dominant spectral components also followed the trends predicted by the above theoretical rotor mechanics model (fig.~\ref{fig:drones_acoustic_signature}C). These measurements established that rotorcraft acoustics contained repeatable physical structure that could be predicted from design variables and observed in real recordings. This observation established the first prerequisite for embodied passive aeroacoustic perception.

\subsubsection*{Ego acoustics and environmental disturbances create the central sensing challenge}

The existence of structured flight acoustics does not imply that relative perception is straightforward. Unlike stationary acoustic sensing systems, an embodied listener additionally operates inside its own aeroacoustic field. Figure~\ref{fig:drones_acoustic_signature}A quantifies how acoustic observability changes as sensing conditions become progressively more realistic. During idle listening, leader-related harmonic structure remained clearly visible. Once the follower began flying, however, its own propellers, motors, and aerodynamic wake introduced substantial ego-acoustic interference. Additional wind further reduced observability. The peak signal-to-noise ratio decreased dramatically across these conditions, indicating that the acoustic information associated with the leader became increasingly sparse relative to the follower’s broadband self-generated sound. This result highlights the central challenge of embodied passive aeroacoustic perception. The problem is not simply detecting drone sound, but recovering weak leader-related signatures embedded within a dynamically changing acoustic field generated by the sensing platform itself.

\subsubsection*{Rotorcraft design influences acoustic observability}

The spectral characterization further reveals an important design principle for embodied passive aeroacoustic perception. Figure~\ref{fig:drones_acoustic_signature}C shows that rotorcraft configurations with different blade counts, drone frames, and payload weights generated distinct harmonic distributions. These observations further suggest that spectral separability between participating rotorcraft can improve acoustic observability. Guided by this principle, we used two otherwise identical UAV platforms differing only in propeller blade count (leader: two blades; follower: three blades), introducing a minimum propulsion-level distinction while preserving nearly identical flight characteristics. As shown in Figure~\ref{fig:drones_acoustic_signature}C, the two-blade leader and three-blade follower exhibited partially separated harmonic structures, producing frequency bands where leader-related signatures could be distinguishable from follower ego acoustics. This effect is further reflected in the power spectral density ratio analysis shown in Figure~\ref{fig:drones_acoustic_signature}D. Several frequency regions exhibited increased leader-to-follower contrast, suggesting that relative perception could benefit from spectral separability between participating platforms. Rather than representing a platform-specific implementation detail, these observations identify a broader design principle: embodied passive aeroacoustic perception becomes increasingly observable when the acoustic signatures of the source and listener occupy at least partially distinguishable spectral regions.

\subsubsection*{Acoustic observability informs sensor design}

The acoustic characterization directly informs our design of the onboard sensing hardware. Figure~\ref{fig:drones_acoustic_signature}E shows that acoustic intensity decreased with range, reducing the available information at larger separations. At the same time, Figure~\ref{fig:drones_acoustic_signature}F shows that the angular resolution achievable by a microphone array improved with frequency. These competing effects create a design trade-off. Lower frequencies propagate farther but provide weaker spatial discrimination, whereas higher frequencies offer stronger directional cues but attenuate more rapidly and become increasingly susceptible to phase ambiguity. We therefore customized our microphone array by selecting the spacing between the MEMS microphones and the type of the MEMS microphone based on these trade-offs to preserve spatial information over the frequency range where multirotor harmonic structure remains both observable and informative. More details are provided in Materials and Methods. Our design provided the physical basis for the sensing architecture used throughout the remainder of the study.

\subsection*{Recovering relative state from passive flight acoustics}

Having established that multirotor flight acoustics contain structured and observable information with principled design choices on the rotorcraft and onboard microphone arrays, we next investigate whether relative state can be recovered from these signals.

\begin{FPfigure}
    \centering
    \includegraphics[width=\linewidth]{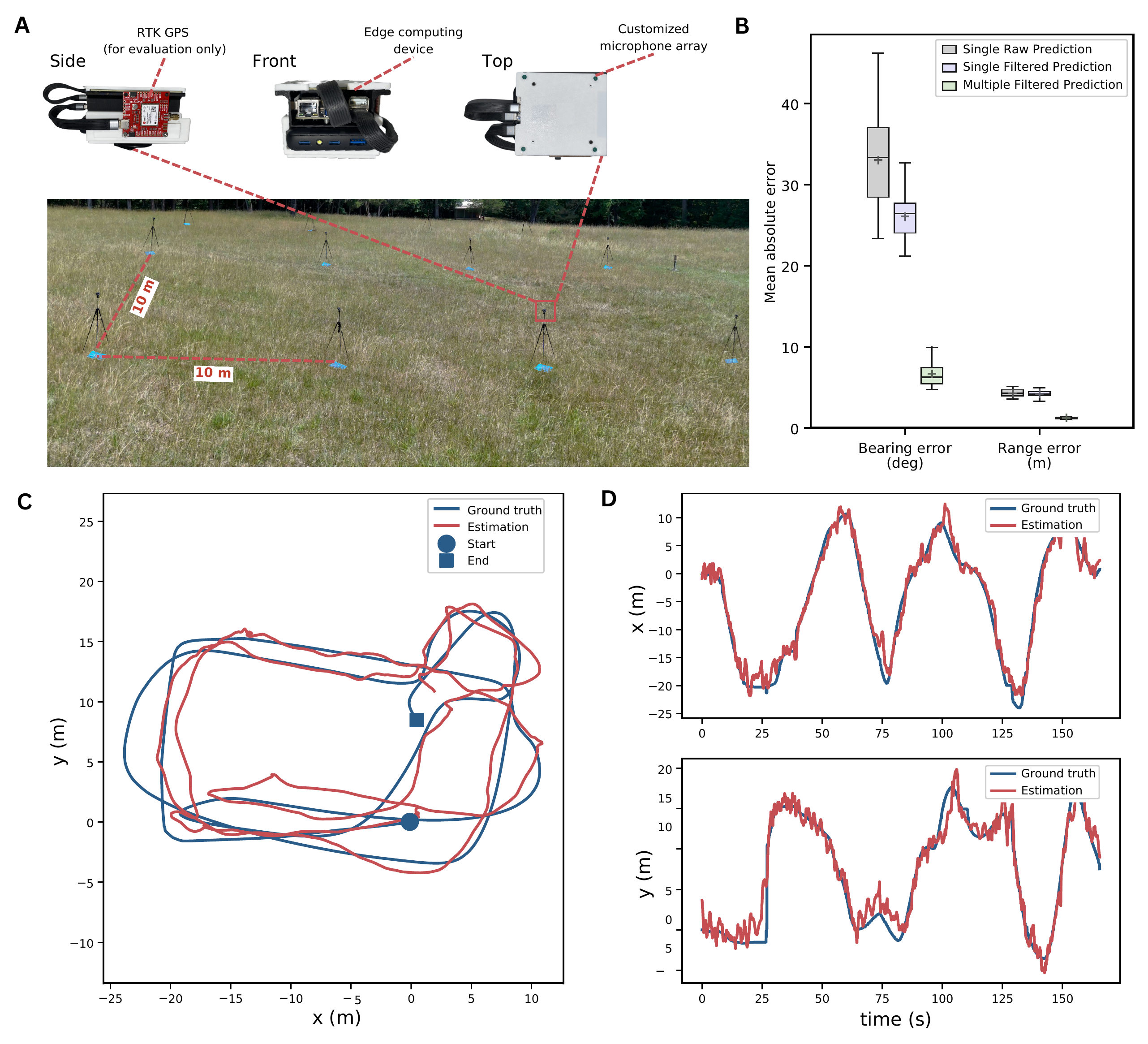}
    \caption{\textbf{Distributed acoustic localization and trajectory reconstruction using a sensor ground unit (SGU) array.}
    (\textbf{A}) Experimental setup showing the distributed SGU deployment over a \SI{30}{\meter} \(\times\) \SI{20}{\meter} outdoor area and the sensing hardware, including a customized microphone array and edge-computing device used for acoustic localization. 
    (\textbf{B}) Open-field trajectory tracking after alignment over \SI{180.06}{\second}, showing ground-truth and estimated leader positions. The ground-truth trajectory spans \SI{282.56}{\meter} with \(n=3{,}129\) segments. The associated localization errors for single-array raw, single-array filtered, and consensus multi-array fusion are summarized by boxplots of \(n=21\) random-bin mean errors (median, interquartile range (IQR), and \(1.5\times\) IQR whiskers). Multi-array fusion achieves the lowest bearing MAE (\SI{6.86}{\degree}) and range RMSE (\SI{1.26}{\meter}).
    (\textbf{C}) Reconstructed two-dimensional target trajectory from acoustic localization estimates (red) compared with GPS ground truth (blue). Circular and square markers denote the start and end positions, respectively. The estimated trajectory closely follows the ground-truth path throughout the experiment.
    (\textbf{D}) Time-series comparison of estimated and ground-truth \(x\)- and \(y\)-coordinates. The acoustic localization estimates accurately capture the temporal dynamics of target motion and exhibit agreement with the GPS reference measurements.}
    \label{fig:static_error}
\end{FPfigure}

\subsubsection*{Distributed listening validates acoustic spatial regression}

Before addressing the substantially more challenging embodied flight setting, we first asked a simpler question: do naturally generated multirotor flight sounds contain sufficient information to recover relative state under controlled but outdoor listening conditions? This experiment isolates the acoustic perception problem from ego-aeroacoustic interference and flight dynamics, providing a proof-of-principle validation that passive flight acoustics alone support spatial regression.

We deployed 12 sensing ground units in a structured \(3 \times 4\) grid with approximately \SI{10}{\meter} spacing in an outdoor field, covering an area of roughly \SI{30}{\meter} \(\times\) \SI{20}{\meter} as shown in Figure~\ref{fig:static_error}A. Each sensing unit included a microphone array, edge computing module, and GPS receiver. During the leader-drone flight, each unit inferred range and bearing from acoustic observations. These local estimates were fused with the sensing-unit GPS positions to reconstruct the leader position and were evaluated against the leader UAV GPS trajectory. We note that the GPS on the sensing unit only served as the anchor for the location of the unit, hence the GPS signals were not part of the neural perception module (see Materials and Methods) for range and bearing prediction.

Figure~\ref{fig:static_error}B compares raw single-array predictions, filtered single-array estimates, and consensus estimates from the distributed sensing array. This comparison separates the contribution of the acoustic regressor, temporal filtering, and spatial diversity across sensing units. The distributed acoustic array achieved a bearing MAE of \SI{6.86}{\degree} and a range RMSE of \SI{1.26}{\meter} over test trajectories with a total length of \SI{282.56}{\meter} (\SI{180.06}{\second}). The reconstructed trajectory in Figure~\ref{fig:static_error}C and the position time series in Figure~\ref{fig:static_error}D show how local bearing-range estimates recover the global leader motion, demonstrating that passive multirotor acoustics contain sufficient information to recover target motion under controlled listening conditions. Our experiments establish that relative state can be inferred from naturally generated flight sound before introducing the substantially more challenging embodied sensing scenario.

Additionally, this experiment also informed us of the relative distance between the leader and follower in our following embodied flight experiment. Specifically, the above results were constrained by the physics of acoustic propagation. As the range increased, the received acoustic intensity decreased according to the inverse-square relationship
\begin{equation}
I(r) = \frac{P}{4\pi r^{2}},
\label{eq:inverse_square}
\end{equation}
where \(I(r)\) denotes received acoustic intensity at distance \(r\), and \(P\) denotes acoustic power emitted by the source. The corresponding reduction in sound pressure and signal-to-noise ratio weakens the spectral and spatial cues available to the regressor. This attenuation sets the baseline difficulty for the onboard case, where the same leader signal must be recovered in the presence of follower ego acoustics and flight-induced disturbances.

While the distributed sensing experiment establishes that naturally generated multirotor aeroacoustics contain sufficient information for spatial regression, it does not yet address the central challenge of embodied passive aeroacoustic perception. In an onboard setting, the sensing platform itself becomes a dominant acoustic source, continuously modifying the acoustic field through its own propellers, motion, and changing geometry relative to the leader. We therefore next investigate whether actionable relative-state information remains recoverable under these substantially more challenging embodied flight conditions.

\subsection*{Onboard listening remains informative during flight}

\subsubsection*{Leader-related cues persist under ego-acoustic interference}

We leveraged the subtle difference caused by rotor dynamics between the 2-blade leader and the 3-blade follower to extract meaningful information about the leader by overcoming follower ego noise. As seen in Fig.~\ref{fig:drones_acoustic_signature}C and D, these rotorcraft differences create frequency-dependent acoustic structure and leader-to-follower spectral contrast. First, the four microphones were placed at a distance of \SI{75}{\milli\meter} to capture a unique time delayed signal. Second, our proposed neural estimator not only took four-channel spectrograms as input but also the IPD and ILD, as summarized in the overall computational pipeline in Figure~\ref{fig:sonicfly_overview}C. The log-magnitude spectrogram features high-amplitude, low-frequency components prominently on the log-frequency scale. This aligns well with the physical reality of acoustic signal propagation in the air and corresponding attenuation, as the low-frequency components from the leader drone would travel further with less attenuation compared to the high-frequency counterparts. Furthermore, IPD highlights the phase difference, assisting in bearing estimation of the leader drone, whereas ILD contributes to range. The contribution of these feature channels was further evaluated in Figure~\ref{fig:comparison}. Coupled with the follower's own position, the relative location of the leader drone could be tracked and pursued.

\begin{figure}
    \centering
    \includegraphics[width=\linewidth]{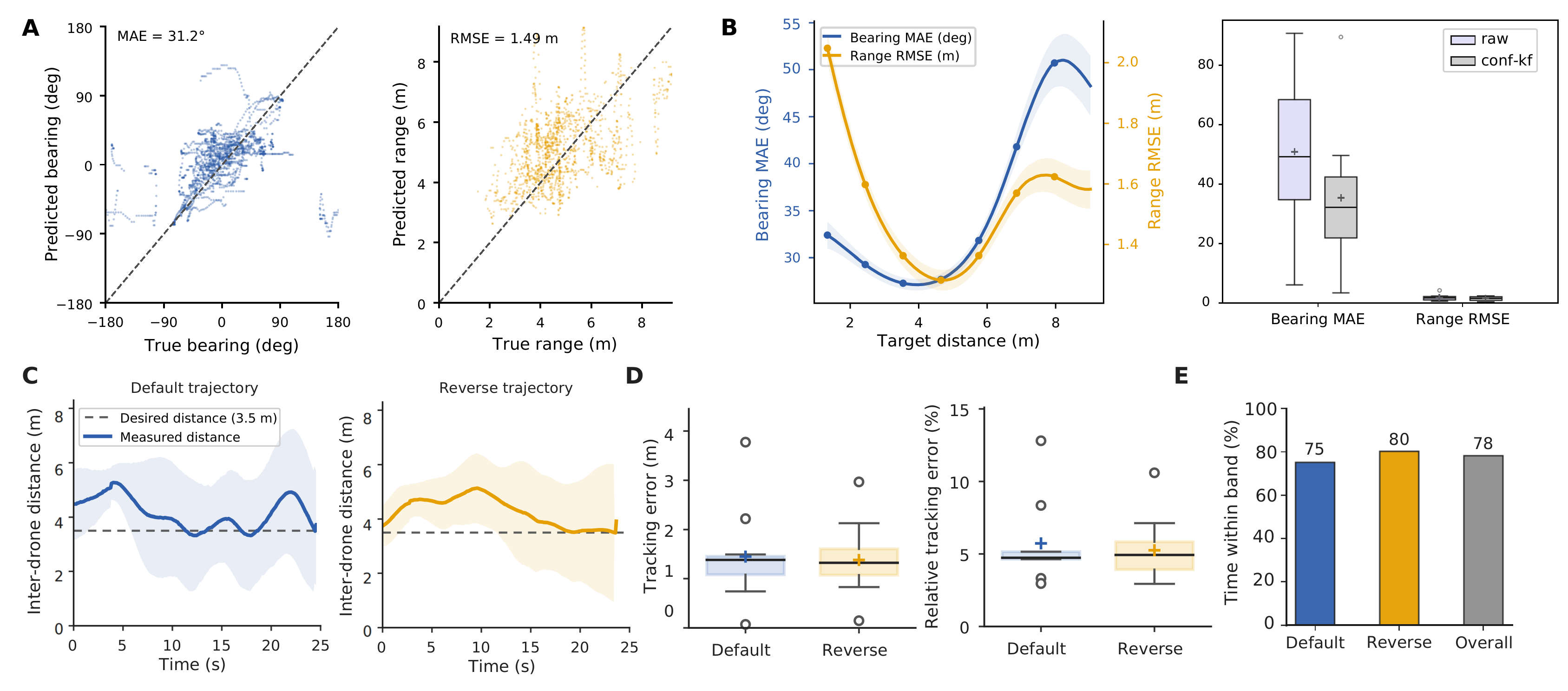}
    \caption{\textbf{Onboard acoustic relative-state estimation and outdoor leader--follower tracking.}
    (\textbf{A}) Offline validation of acoustic relative-state estimates against RTK-derived ground truth. Scatter plots compare predicted and true bearing and range measurements after confidence-gated Kalman filtering (Conf-KF).
    (\textbf{B}) Localization performance as a function of inter-drone distance. Bearing mean absolute error (MAE) and range root mean square error (RMSE) are lowest at intermediate separations. Field localization error comparison between single-array raw predictions and Conf-KF predictions across \(n=10\) trajectories (\(N=2{,}296\) paired samples; total trajectory length \SI{334.84}{\meter}). Bearing and range errors are summarized per trajectory as MAE and RMSE, respectively, using boxplots (IQR boxes, median lines, mean crosses, and whiskers extending to \(1.5\times\) IQR). Conf-KF improves localization accuracy, reducing mean bearing MAE from \SI{50.83}{\degree} to \SI{35.34}{\degree} and mean range RMSE from \SI{1.82}{\meter} to \SI{1.48}{\meter}.
    % (\textbf{B}) Localization accuracy is qualitatively highest at intermediate inter-drone separations. 
    (\textbf{C}) Inter-drone distance during leader--follower flights under default and reverse trajectory conditions. The dashed line indicates the desired separation distance of \SI{3.5}{\meter}, and shaded regions denote variability across repeated trials.
    (\textbf{D}) Absolute and relative tracking error distributions for default and reverse flight conditions.
    (\textbf{E}) Percentage of flight time spent within the desired tracking band. The follower remained within the target band for \SI{75}{\percent} of default trials, \SI{80}{\percent} of reverse trials, and \SI{78}{\percent} overall.}
    \label{fig:onboard_acoustic_performance}
\end{figure}

\subsubsection*{Onboard acoustic relative-state estimation during flight}

We next evaluated whether onboard acoustic measurements retain relative-state information over the follower's strong ego noise during outdoor flight. In this embodied setting, the array was exposed to follower ego-motion, propeller self-noise, wind, attitude variation, and continuously changing source-listener geometry. Raw predictions and filtered aeroacoustic state estimates from our neural estimators were evaluated on held-out onboard flight data and compared with time-aligned Real-Time Kinematic (RTK) GPS-derived relative pose.

Figure~\ref{fig:onboard_acoustic_performance}A summarizes onboard acoustic localization performance, with the post-processing algorithm described in the Materials and Methods subsection ``Robust tracker on inferences''. Scatter plots compare predicted bearing and range against RTK ground truth. Despite strong ego-acoustic interference, continuously changing geometry, and environmental disturbances, acoustic measurements remained predictive of relative state. After confidence-gated filtering, SonicFly achieved a bearing MAE of \SI{31.2}{\degree} and a range RMSE of \SI{1.49}{\meter} on held-out flight data.

We further quantified distance-dependent performance in Figure~\ref{fig:onboard_acoustic_performance}B. Interestingly, localization error exhibited a non-monotonic dependence on leader distance. Bearing MAE and range RMSE were both lowest at intermediate distances. These results showed that closer proximity between the two drones does not always benefit the maintenance of team structure. At short range, the aeroacoustics of the leader drone may be dominated by near-field multirotor structure, wake interactions, and rapid geometry-dependent changes in the follower's received signals. At long range, attenuation can reduce leader-related harmonic energy relative to the background noise and ego noise, which can weaken both spectral and spatial cues. Figure~\ref{fig:onboard_acoustic_performance}B shows that raw predictions yielded a bearing mean absolute error (MAE) of \SI{45.7}{\degree} and a range root mean square error (RMSE) of \SI{1.68}{\meter}. By explicitly considering prediction uncertainty and smoothing, confidence gating followed by Kalman filtering reduced these errors to a bearing MAE of \SI{31.2}{\degree} and a range RMSE of \SI{1.49}{\meter}. Our neural estimator's confidence provided a useful criterion for rejecting low-quality estimates and recovering a more reliable relative-state signal from onboard acoustic measurements. Our results show that actionable relative-state information can be reasonably retrieved even under embodied flight conditions, which provides the basis of leveraging this information for closed-loop aerial pursuit.

\subsection*{Passive aeroacoustic perception supports closed-loop outdoor aerial pursuit}
% \el{Figure~\ref{fig:onboard_acoustic_performance}C shows leader and follower trajectories overlaid across repeated trials in both trajectory conditions. The follower trajectory captured the qualitative motion structure of the leader in both default and reverse conditions, indicating that the onboard acoustic estimates remain consistent with the relative motion observed in the field. Deviations were prominent near turns and transient segments, where acoustic ambiguity, estimator latency, and vehicle dynamics can compound tracking error.}

Recovering relative state alone does not establish embodied passive aeroacoustic perception. Because every control action changes subsequent acoustic observations, the ultimate test is whether the recovered information is sufficiently actionable to support stable closed-loop flight. We therefore deployed SonicFly in outdoor leader-follower pursuit experiments where the objective for the follower drone was to maintain a desired separation distance with onboard acoustic relative-state estimates.

As shown in Figure~\ref{fig:onboard_acoustic_performance}C, the inter-drone distance remained centered around the nominal \SI{3.5}{\meter} target spacing under both default and reverse ``sine'' trajectory conditions. The follower drone was only trained using data collected with the default ``sine'' trajectories. Though the test data may still follow the default patterns, the environmental conditions and follower-leader geometry still varied from time to time. Therefore, the default ``sine'' trajectories evaluated the interpolation capability. On the other hand, the reverse ``sine'' trajectories followed the opposite direction than those in the training data, serving as test data for generalization. As shown in the results, while the separation distance exhibited time-varying oscillations arising from tracking dynamics, estimation noise, and environmental disturbances, the follower consistently maintained the desired field-scale spacing throughout the experiment. The shaded regions indicate variability across repeated trials and demonstrate the repeatability of the tracking behavior.

To further quantify tracking performance, Figure~\ref{fig:onboard_acoustic_performance}D presents the distributions of both absolute tracking error and relative tracking error for the two trajectory conditions (i.e., default and reverse ``sine'' trajectories). Additionally, we defined a \emph{within-band} metric that measures whether the follower remains within a lateral corridor centered on the leader trajectory and bounded by a predefined threshold (details are in Materials and Methods). As summarized in Figure~\ref{fig:onboard_acoustic_performance}E, the follower remained within the desired lateral band, defined as the leader position \(\pm\)~\SI{3.5}{\meter}, for \SI{75}{\percent} of the default trajectory duration, \SI{80}{\percent} of the reverse trajectory duration, and \SI{78}{\percent} overall. The results indicate comparable performance across experiments and confirm that passive aeroacoustic perception provided sufficient information to guide dynamic embodied decision making and sustain closed-loop aerial pursuit.

To evaluate the generalization capability of SonicFly, we conducted additional field experiments spanning more than twenty leader trajectories, including slanted straight lines, C-shaped curves, sigmoidal paths, S-shaped trajectories, L- and inverse-L-shaped turns, \(\Gamma\) and inverse-\(\Gamma\)-shaped trajectories, and trajectory patterns spelling ``grL'' and ``DUKE''. Representative examples are shown in Figure~\ref{fig:sonicfly_overview} and Figure~\ref{fig:field_trajectories}. Across these diverse motion profiles, the follower consistently tracked the leader and maintained relative proximity using only onboard aeroacoustic perception, demonstrating robustness to varying path geometries, turning behaviors, and trajectory complexities. Our results show that embodied passive aeroacoustic perception generalizes beyond a single trajectory pattern and supports a broad range of outdoor aerial pursuit behaviors.

\begin{FPfigure}
    \centering
    \includegraphics[width=\linewidth]{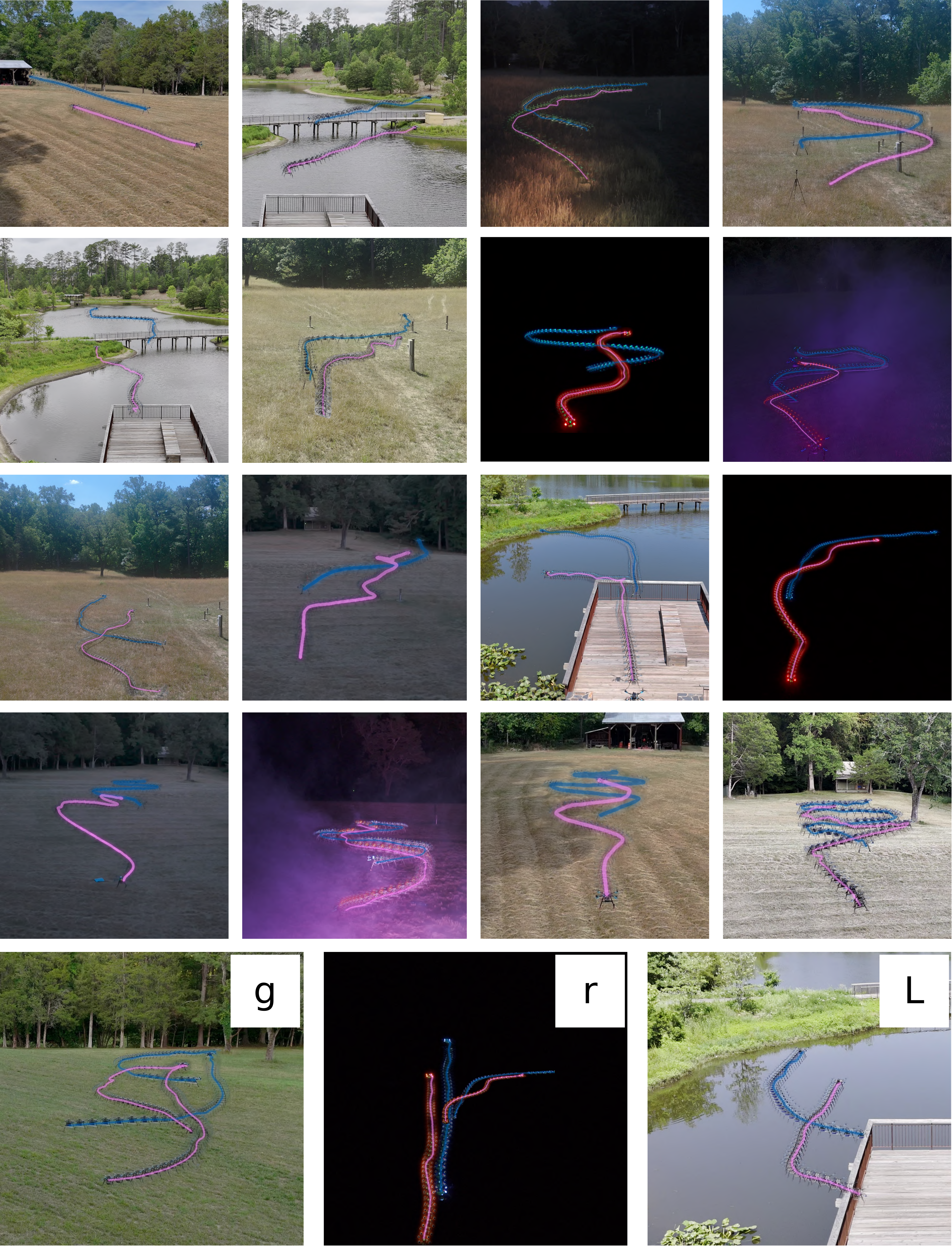}
    \caption{\textbf{Outdoor acoustic pursuit trajectory montage.}
    Trajectory-level visualization of field pursuit runs across diverse path geometries and environmental conditions. Blue and magenta traces denote the leader and follower trajectories, respectively.}
    \label{fig:field_trajectories}
\end{FPfigure}

\begin{FPfigure}
    \centering
    \includegraphics[width=\linewidth]{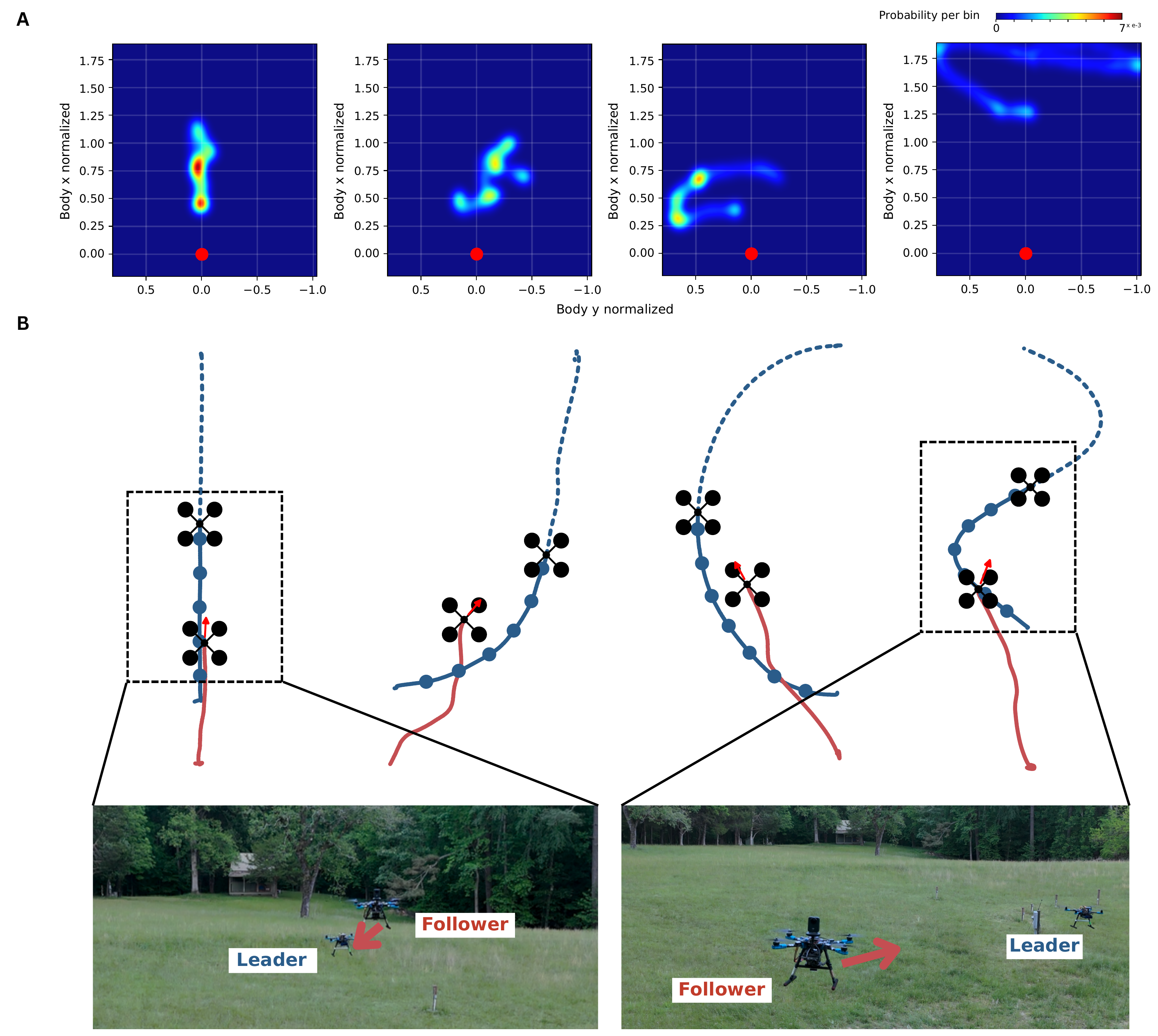}
    \caption{\textbf{Closer examination of acoustic-guided pursuit behavior.}
    (\textbf{A}) Distribution of leader positions expressed in the follower body frame for representative pursuit trajectories. The red marker denotes the follower location, and the heatmap shows the accumulated relative positions of the leader during tracking. Across diverse trajectory classes, the leader remains concentrated within the forward sector of the follower body frame, indicating successful pursuit and maintenance of target observability.
    (\textbf{B}) Representative outdoor pursuit snapshots. Blue and red trajectories denote the leader and follower UAVs, respectively, and black markers indicate sampled intermediate positions. Insets show corresponding field images. The follower continuously adjusts its trajectory to reduce relative displacement and maintain pursuit across trajectories with varying curvature and turning behaviors.}
    \label{fig:microscope_trajs}
\end{FPfigure}

\begin{FPfigure}
    \centering
    \includegraphics[width=\linewidth]{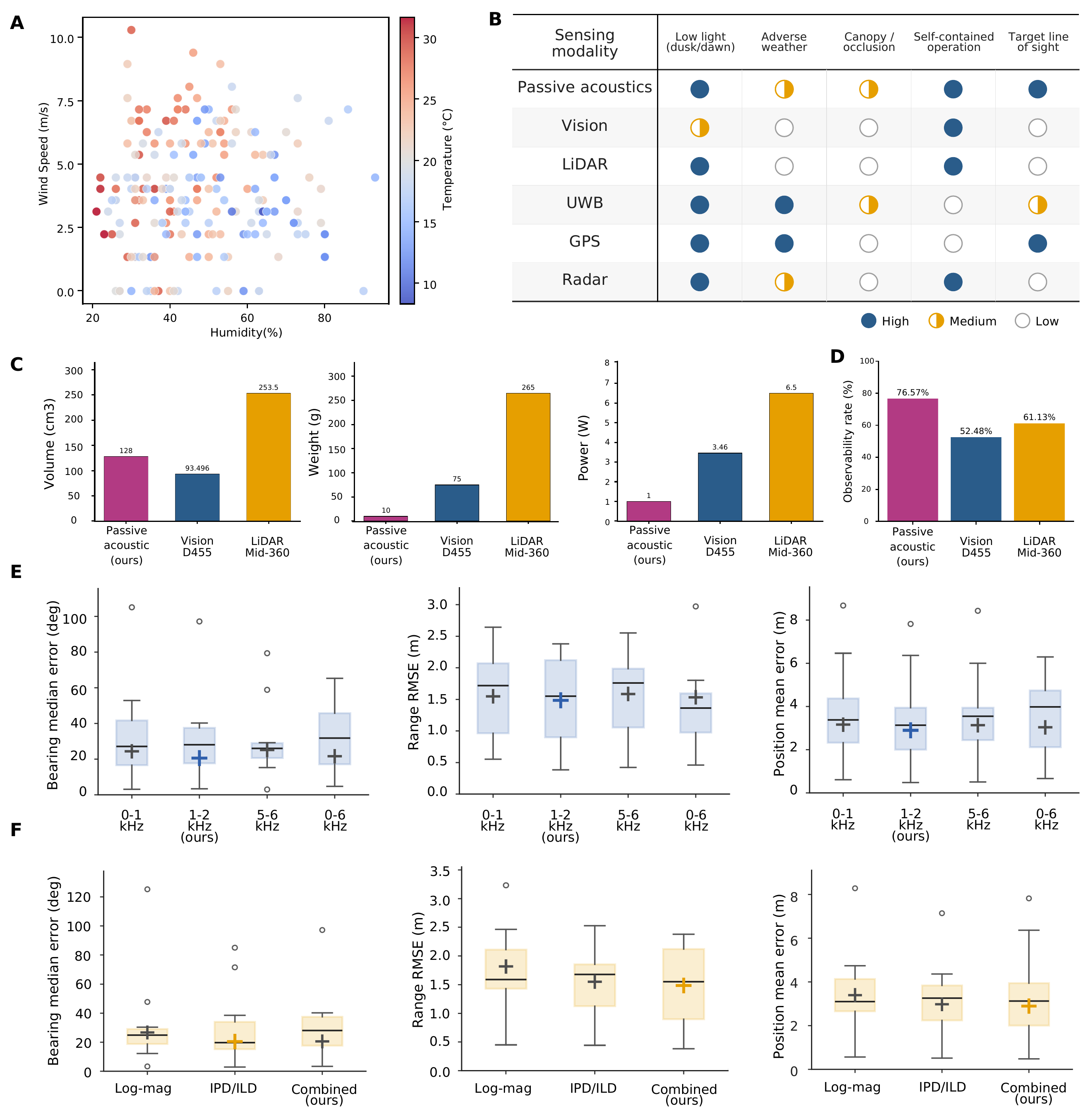}
    \caption{\textbf{Robustness across field conditions and degraded sensing scenarios.}
    (\textbf{A}) Environmental distribution of outdoor SonicFly deployments, showing wind speed, humidity, and temperature across flight sessions.
    (\textbf{B}) Comparison of sensing modalities for aerial relative tracking across representative operating constraints. Filled, half-filled, and open circles indicate high, medium, and low suitability, respectively.
    (\textbf{C}) Size, weight, and power data of the custom acoustic payload compared with D455 RealSense camera, and Livox Mid-360 LiDAR sensors.
    (\textbf{D}) Observability-rate comparison under degraded sensing conditions, including vision and LiDAR modalities.
    (\textbf{E}) Frequency-band ablation shows that the \SIrange{1}{2}{\kilo\hertz} spectrogram provides the best overall trade-off, with the lowest bearing error and competitive range and position errors. 
    (\textbf{F}) Feature-channel ablation compares log-magnitude, IPD/ILD, and combined inputs within the selected \SIrange{1}{2}{\kilo\hertz} band. IPD/ILD cues carry most bearing information, whereas combining IPD/ILD with log-magnitude cues improves overall position accuracy. 
Boxes show interquartile ranges, center lines show medians, plus signs show means, and points show outlier trials.}
    \label{fig:comparison}
\end{FPfigure}

A closer examination of the pursuit behavior is shown in Figure~\ref{fig:microscope_trajs}. Figure~\ref{fig:microscope_trajs}A visualizes the spatial distribution of leader locations expressed in the follower's body frame across several representative trajectory classes. The red marker denotes the follower position, while the heatmap represents the accumulated relative positions of the leader observed during successful pursuit. Despite substantial differences in trajectory geometry, the leader consistently remained concentrated within the forward sector of the follower body frame, indicating that the follower actively maintained acoustic observability while pursuing the target. Figure~\ref{fig:microscope_trajs}B presents representative pursuit snapshots from outdoor experiments. The follower adapted its trajectory in response to the evolving relative geometry, generating corrective maneuvers that steered it toward the leader and maintained pursuit across a range of curved and turning trajectories.

Our real-world field experiments demonstrate that embodied passive aeroacoustic perception provides actionable relative-state information under outdoor flight conditions. Rather than relying on external microphone infrastructure, GPS-based relative positioning, or inter-robot communication, the follower used only the leader's intrinsic flight sound to guide pursuit. The successful tracking of a wide range of trajectory shapes further suggests that acoustic sensing is a viable and generalizable modality for relative perception and navigation between aerial robots operating in dynamic outdoor environments.

\subsection*{Robustness across deployment conditions}

As outdoor acoustic sensing is inherently influenced by atmospheric conditions and environmental variability, such as sound propagation speed, attenuation, turbulence, and changing ambient noise, we characterized the environmental conditions under which SonicFly was evaluated. Unlike controlled laboratory experiments, open-field robotic deployments expose the acoustic channel to wind, humidity, temperature variation, precipitation history, and changing ambient sound.

Across our experiments, the collected weather statistics exhibited substantial variability in atmospheric and acoustic propagation conditions (fig.~\ref{fig:comparison}A). In particular, ambient temperatures ranged from approximately \SI{8}{\degreeCelsius} to \SI{32}{\degreeCelsius}, relative humidity between \SI{19}{\percent} and \SI{93}{\percent}, and wind speeds ranged from near-zero conditions to approximately \SI{37}{\kilo\meter\per\hour}. The dataset also contained diverse environmental and illumination conditions, including clear skies, scattered and broken cloud coverage, overcast conditions, and post precipitation. Notably, several deployment sessions were conducted under high-wind and high-humidity conditions, which are known to affect outdoor sound propagation through wind-induced refraction, turbulence, and humidity-dependent atmospheric absorption \cite{piercy1977review,bass1995atmospheric}. The successful operation of SonicFly across these conditions suggests that embodied passive aeroacoustic perception remains viable under realistic outdoor disturbances and environmental variability.

\subsection*{Positioning embodied passive aeroacoustics within the aerial sensing landscape}

Embodied passive aeroacoustic perception is not intended to replace widely used sensing techniques such as cameras, LiDAR, radar, GPS, or communication-based localization. Instead, it occupies a complementary operating regime. Figure~\ref{fig:comparison}B summarizes the strengths and limitations of several sensing modalities. Compared with vision and LiDAR systems, passive acoustics provides a lightweight, self-contained, and complementary sensing modality that remains functional under low-light conditions and does not require direct visual observation of the target. Figure~\ref{fig:comparison}C further shows that the acoustic payload achieves substantially lower size, weight and power consumption (SWaP) than other representative vision and LiDAR alternatives. Specifically, our SonicFly acoustic payload occupies approximately \SI{128}{\centi\meter\cubed}, weighs \SI{10}{\gram}, and consumes \SI{1}{\watt}, compared with \SI{93.5}{\centi\meter\cubed}, \SI{75}{\gram}, and \SI{3.46}{\watt} for the Vision D455, and \SI{251.3}{\centi\meter\cubed}, \SI{265}{\gram}, and \SI{6.5}{\watt} for the LiDAR Mid-360. While the acoustic system occupies slightly more volume than the vision sensor, it achieves substantially lower weight and power consumption.

To assess practical observability, Figure~\ref{fig:comparison}D reports the observability rate, defined as the percentage of time for which the target leader drone was successfully detected and the resulting relative-state estimate satisfied a predefined translation-error threshold (\SI{5}{\meter}). All acoustics, vision and LiDAR systems were evaluated on the same dataset collected under challenging environmental conditions as discussed above in our acoustics experiments, including evening lighting and haze. Under these conditions, the acoustic-based system reached \SI{76.57}{\percent}, vision-based system achieved an observability rate of \SI{52.48}{\percent}, and the LiDAR-based system achieved \SI{61.13}{\percent}. While passive acoustics typically does not provide the precision of these modalities under ideal conditions, it offers a fundamentally different information channel that can complement existing sensing approaches when visibility, infrastructure, or communication become unreliable.

\subsection*{Embodied passive aeroacoustic perception emerges from complementary spectral and spatial information}

Our results so far have established that actionable relative-state information remains recoverable from embodied passive aeroacoustic measurements despite strong ego-acoustic interference and environmental variability. We here investigate which components of the acoustic signal contribute most strongly to this capability. Rather than treating SonicFly as a purely data-driven system, we seek to identify the physical and algorithmic factors that make embodied passive aeroacoustic perception possible.

\subsubsection*{Not all acoustic frequencies contribute equally to relative perception}

We first investigated whether localization performance depends was driven primarily by acoustic energy or by spatially discriminative spectral structure. Based on the aperture of our selected microphones, we limited the broadband input to \SIrange{0}{6}{\kilo\hertz} to reduce the contribution of frequency regions where spatial aliasing and microphone-to-microphone response variability are more likely. Within this range, Figure~\ref{fig:comparison}E compares models trained using four frequency ranges spanning low-frequency rotor energy (\SIrange{0}{1}{\kilo\hertz}), intermediate-frequency leader structure (\SIrange{1}{2}{\kilo\hertz} ), higher-frequency spatial resolution (\SIrange{5}{6}{\kilo\hertz}), and a broadband representation (\SIrange{0}{6}{\kilo\hertz}).

Interestingly, the best overall performance occurred in the \SIrange{1}{2}{\kilo\hertz}  band, which achieved the lowest post-filtered bearing error and low range error. This band coincided with prominent leader-related spectral structure in the power spectral density (PSD)-ratio analysis in Figure~\ref{fig:drones_acoustic_signature}D, while providing shorter wavelengths than the rotor-dominated low-frequency band. In comparison, the \SIrange{0}{1}{\kilo\hertz} band contained strong rotor and blade-passing energy, but its long wavelengths produced weak inter-microphone phase and level differences for the compact array. The \SIrange{5}{6}{\kilo\hertz} band provided shorter wavelengths and finer nominal spatial resolution, but its higher-order harmonic content tended to be weaker and less stable, and was more sensitive to phase wrapping, multi-path, occlusion, and microphone-response variation. Combined with the distance-dependent reduction in acoustic intensity shown in Figure~\ref{fig:drones_acoustic_signature}E, these effects reduced the actionable information available at higher frequencies.

The broadband \SIrange{0}{6}{\kilo\hertz} input did not improve performance relative to the selected \SIrange{1}{2}{\kilo\hertz} band. This suggests that embodied passive aeroacoustic perception is not simply a matter of incorporating more acoustic information. Instead, low-frequency components introduce strong but weakly directional energy, whereas higher-frequency components contribute spatial cues that are less stable under realistic flight conditions. Restricting the representation to the \SIrange{1}{2}{\kilo\hertz} band therefore acts as an acoustic regularizer, preserving robust multirotor harmonic structure while suppressing frequencies that are either spatially under-resolved or acoustically unstable. We therefore selected the \SIrange{1}{2}{\kilo\hertz} spectrogram as the input representation for our final SonicFly system.

\subsubsection*{Spatial acoustic cues dominate directional perception}

We next investigated which feature channels carried the localization information within the selected \SIrange{1}{2}{\kilo\hertz} representation. Figure~\ref{fig:comparison}F compared three models trained using different acoustic inputs: a log-magnitude-only representation containing per-microphone spectral energy, an IPD/ILD-only representation containing inter-microphone phase and level differences, and a combined representation incorporating both spectral and spatial cues. This comparison separates platform-dependent acoustic signatures from explicit multi-microphone spatial information.

The results reveal that spectral magnitude alone is insufficient for robust localization. The log-magnitude-only model produced larger bearing, range, and position errors, indicating that acoustic energy and platform spectral signatures do not fully determine the leader position relative to the follower. In contrast, the IPD/ILD-only model recovered much of the directional performance, consistent with inter-microphone phase and level differences carrying the dominant bearing information. This observation is consistent with classical acoustic localization theory, in which relative phase and level differences provide directional cues about source location \cite{rayleigh1907perception, middlebrooks1991sound, macpherson2002listener}.

However, the combined representation achieved the lowest overall position and range errors, showing that spectral and spatial information play complementary roles. While IPD and ILD provide the primary directional cues, log-magnitude features contribute additional information about source strength, harmonic structure, and platform-specific acoustic identity. These results suggest that embodied passive aeroacoustic perception emerges from the interaction of two information channels: acoustic signatures that characterize the source and spatial cues that reveal its relative geometry. Our final SonicFly representation therefore combined both components to maximize localization performance.

\section*{Discussion}

This work introduces and demonstrates \textit{embodied passive aeroacoustic perception}, a sensing paradigm in which an aerial robot infers relative-state information from the naturally generated sound of flight and uses this as an actionable signal for aerial pursuit. While previous acoustic localization systems have demonstrated the feasibility of drone-to-drone acoustic sensing with an off-engine drone, indoor environments, or active probing, SonicFly establishes that naturally generated flight sounds remain sufficiently informative for onboard relative perception during simultaneous flight despite strong ego-acoustic interference, environmental variability, and continuously changing source-receiver geometry.. Through acoustic characterization, onboard localization, and closed-loop pursuit experiments in open fields, we demonstrate that multirotor aeroacoustic fields provide an actionable sensing channel that complements existing aerial perception modalities.

More broadly, our results suggest a complementary perspective on how robotic sensing systems could potentially be designed. Across robotics, many physical signals generated as byproducts of robot motion are traditionally treated as disturbances, noise sources, or unwanted byproducts. Examples include acoustic emissions, wake fields, vibrations, thermal signatures, airflow disturbances, chemical particles, and other forms of energy exchanged with the environment. In biological systems, however, similar signals have been discovered as frequent information channels that support coordination, navigation, and collective behavior. Our results presented here suggest that robotic systems may likewise exploit naturally generated signals from motions as sources of information rather than suppressing or ignoring them. From this perspective, embodied passive aeroacoustic perception represents one instance of a broader class of sensing paradigms that leverage the physical consequences of motion itself as actionable information.

Our experiments also provide several insights into the conditions that enable passive aeroacoustic perception to be possible. First, acoustic observability is strongly influenced by rotorcraft configuration. Differences in blade count and propulsion characteristics create partially separable harmonic structures that improve leader-follower contrast and facilitate localization. Second, not all acoustic information contributes equally to relative perception. Intermediate-frequency harmonics provide a favorable balance between acoustic observability and spatial discriminability, whereas lower frequencies contain strong but weakly directional energy, and higher frequencies become increasingly susceptible to attenuation and instability. Third, localization performance emerges from the combination of spectral and spatial information. Acoustic signatures alone are insufficient for robot localization, whereas inter-microphone phase and level differences provide the dominant directional cues. These observations provide design principles for future embodied passive aeroacoustic systems and highlight the importance of jointly considering the ``full-stack'' from platform mechanics, to sensing hardware, and signal representations.

In our closed loop pursuit task, what makes the setting more challenging is the tight coupling between perception and action. Every action taken by the follower UAV changes the future acoustic observations by altering source-receiver geometry, acoustic interference, and target observability. The act of flight therefore creates the signal, perturbs the signal, and determines what we will be sensed next. This coupling suggests that embodied passive perception should be viewed not only as a localization problem, but also as a process in which action and perception are inherently intertwined, since future observations depend directly on the robot's behavior. The pursuit experiments presented here provide an initial demonstration of this concept by showing that passive aeroacoustic estimates can support closed-loop aerial behavior.

An important observation from this study is that embodied passive aeroacoustic perception becomes feasible not because drone sounds are inherently easy to interpret, but because the sensing problem can be jointly shaped through platform design, sensing hardware, and learning. Rotor configuration influences spectral observability, microphone geometry determines the available spatial information, and learned representations recover weak cues that would be difficult to isolate using hand-crafted signal processing alone. Embodied passive aeroacoustic perception therefore emerges from the interaction between mechanics, acoustics, sensing, and computation rather than from any single component in isolation.

Several areas for improvement remain. Acoustic observability is fundamentally constrained by the physics of sound propagation. Leader-related acoustic signals attenuate with distance, higher-order harmonics become weaker and less reliable, and environmental disturbances such as wind introduce additional uncertainty. Because perception and action are tightly coupled, localization errors can directly influence future observations and actions and compound over time. Our current system therefore demonstrates pursuit and relative-state maintenance rather than precise formation control. Furthermore, the present work focuses on leader-follower interactions involving a single target and observer pair. More complex scenarios involving multiple robots, dense environments, or highly dynamic maneuvers remain important open directions for future studies.

Meanwhile, several opportunities emerge from this sensing paradigm. Future systems may actively exploit acoustic observability during planning and control, selecting actions that improve future measurements rather than treating sensing as a passive process. Multi-agent formulations may allow teams of robots to jointly localize, communicate, or coordinate through naturally generated acoustic traces. More generally, SonicFly raises the possibility of robotic systems that perceive and coordinate through the physical consequences of their own motion, extending beyond acoustics to other forms of environmental interaction. Therefore, we view SonicFly not as a replacement of existing sensing modalities, but as an initial demonstration of a broader sensing paradigm in which naturally generated signals from motions become useful sources of information for robotic perception and coordination.

\section*{Materials and Methods}

\subsection*{UAV platforms and embodied acoustic sensing system}

\subsubsection*{UAV platforms and RTK reference}

The leader and follower UAVs were built on Holybro X500 V2 quadrotor platforms equipped with Pixhawk 6X flight controllers and H-RTK F9P RTK positioning modules. The follower used its onboard state estimator for flight stabilization and trajectory execution. However, no leader GPS, RTK, communication, or externally referenced relative-position information was provided to the follower's neural estimator, the confidence-gating module, or the EKF. RTK measurements were used only for dataset labeling and post hoc evaluation.

The leader vehicle retained the baseline Holybro propulsion configuration, whereas the follower vehicle was upgraded with BLHeli\_32 electronic speed controllers and BrotherHobby 2812 \SI{900}{kV} motors to accommodate the additional sensing and computing payload. To improve acoustic observability under strong follower ego-noise, the two vehicles employed different propeller configurations. The leader used two-blade propellers, whereas the follower used three-blade propellers, producing related but partially separable harmonic structures. This design was motivated by our investigation of the acoustic conditions that enable embodied passive aeroacoustic perception. As shown in the Results section, rotorcraft configuration influences the spectral separability between leader and follower acoustics, thereby affecting the amount of actionable information available for relative-state estimation. The selected two-blade/three-blade configuration provided a practical operating regime in which leader-related harmonic signatures remained observable despite strong follower ego-acoustic interference.

\begin{figure}
    \centering
    \includegraphics[width=\linewidth]{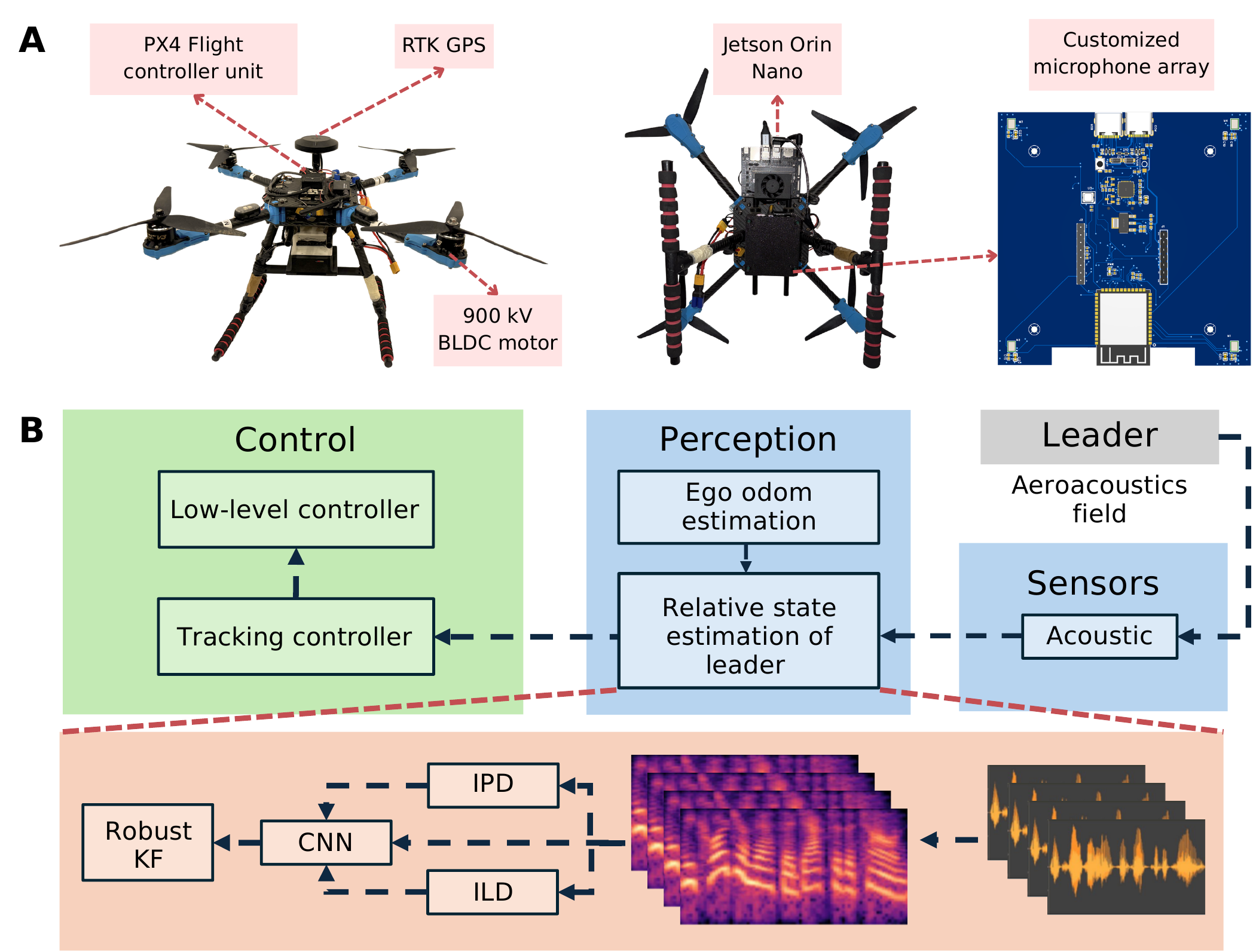}
    \caption{\textbf{SonicFly hardware platform and aeroacoustic pursuit architecture.}
    (\textbf{A}) Custom quadrotor aerial platform used for SonicFly deployments, showing integration of the PX4 flight controller, RTK-GPS module, and lightweight MEMS microphone-array acoustic sensing payload.
     (\textbf{B}) System-level architecture of the acoustic leader-tracking pipeline. Acoustic signals generated by the leader platform are transformed into spectrogram, IPD, and ILD features, followed by a convolutional neural network (CNN)-based perception and robust Kalman filtering to estimate the relative state of the leader for closed-loop follower control.}
    \label{fig:archi}
\end{figure}

The follower UAV carried our customized four-microphone array as the acoustic payload together with a Jetson Orin Nano for onboard inference and pursuit control. Our microphone array consisted of four microphones mounted on a square printed circuit board with \SI{75}{\milli\meter} inter-microphone spacing, further described in the following sections. Additional platform details are shown in Figure~\ref{fig:archi}A.

\subsubsection*{Design principles of the embodied acoustic sensor} \label{sec:acoustic_sensor_design}

Embodied passive aeroacoustic perception requires a sensing system that can simultaneously satisfy the constraints of aerial deployment and acoustic localization. The sensor must be lightweight, synchronized across channels, capable of measuring spatial acoustic cues, and compatible with the limited payload capacity of a multirotor platform. Existing microphones are often bulky or only include one or two microphones, lacking the ability to capture three-dimensional acoustic signatures. Other microphone arrays are mostly designed to optimize for indoor recording such as conference rooms. To satisfy the requirements of embodied passive aeroacoustic perception, we developed a customized lightweight four-channel microphone array based on Micro-Electro-Mechanical Systems (MEMS) microphones. MEMS microphones provide miniature silicon-etched acoustic sensors that achieve a favorable balance between sensing performance and size, weight, and power (SWaP) requirements compared with conventional condenser microphones, albeit with comparatively lower SNR performance~\cite{ghenescu2026acoustic}. Our sensor array employs four TDK InvenSense ICS-43434 digital MEMS microphones \cite{ics43434_datasheet} arranged symmetrically on an \SI{83.5}{\milli\meter} $\times$ \SI{83.5}{\milli\meter} printed circuit board and synchronized through an ESP32-S3-WROOM-1 microcontroller \cite{esp32s3_datasheet}. The microphones provide an SNR of approximately $65$~dBA (A-weighted decibels) and a sensitivity of $-26$~dBFS (decibels relative to full scale).

The array was mounted beneath the follower UAV and positioned as far as practical from the propulsion motors to reduce rotor-induced airflow interference while preserving an unobstructed acoustic path toward the leader. The array geometry was selected to preserve useful inter-microphone phase and level differences while minimizing excessive spatial aliasing. For a microphone spacing \(d\), the half-wavelength spatial-aliasing condition is
\begin{equation}
\label{eq:spatial_alias}
d < \frac{c}{2 f_{\max}}
\end{equation}
where \(c\) denotes the speed of sound in air and \(f_{\max}\) represents the highest unambiguous frequency. With an adjacent microphone spacing of \(d = \SI{0.075}{\meter}\) and \(c = \SI{343}{\meter\per\second}\), the adjacent-pair spatial-aliasing limit is approximately \SI{2287}{\hertz}. This range overlaps the frequency bands identified in the Results section as containing informative leader-related harmonic structure.

The achievable angular resolution of a microphone array scales with both acoustic wavelength and effective aperture,
\begin{equation}
\label{eq:angular_resolution}
\theta_{\text{res}} = \frac{\lambda}{N d}
= \frac{c / f_0}{N d}
\end{equation}
where \(\theta_{\text{res}}\) is the minimum resolvable angular separation, \(\lambda\) is acoustic wavelength, \(N\) is the number of microphones, and \(f_0\) is the frequency of interest. The corresponding angular resolution trend as a function of frequency has been shown earlier in Figure~\ref{fig:drones_acoustic_signature}F. Consequently, the array geometry reflects a trade-off between spatial resolution and aliasing robustness, which motivated our frequency-band selection discussed in the Results section.

Acoustic data were acquired at \SI{12000}{\hertz} through synchronized Inter-IC Sound (\(I^2S\)) interface using two stereo \(I^2S\) ports on the ESP32 microcontroller. Signal processing was performed in two stages across the ESP32 and an onboard NVIDIA Jetson Orin Nano. On the ESP32, microphone data were converted into a usable 32-bit format, downsampled to 16-bit payloads, and streamed over native USB, after which the messages were communicated through the Robot Operating System 2 (ROS 2) protocol. On the Jetson platform, the received acoustic signals were normalized and transformed into mel-scale spectrograms for subsequent feature extraction and inference.

\subsection*{Flight dataset formation}

Training and evaluation data were collected through outdoor paired-flight experiments involving a leader and follower UAV. The leader UAV was commanded at \SI{20}{\hertz} to follow predefined planar trajectory profiles, including curved and reversed paths that varied the relative bearing, range, and lateral motion observed by the listener. During primary data collection, the follower was controlled by a pose-based Proportional-Derivative (PD) controller at \SI{20}{\hertz}, using RTK-derived relative pose rather than acoustic feedback, ensuring stable coverage of the desired operating envelope while maintaining independence between data collection and learned acoustic inferences. The follower was commanded to maintain horizontal separations ranging from \SI{1.5}{\meter} to \SI{7}{\meter}. To reduce direct rotor-wake interaction aerodynamic interaction between the UAVs, the follower was flown approximately \SI{0.5}{\meter} above the leader during paired-flight data collection. We used DAgger (Dataset Aggregation)-inspired data collection \cite{ross2011reduction} to capture rare acoustic configurations and failure cases that were difficult to obtain through nominal pursuit trajectories alone, such as jittering or flying in the opposite direction.

For each flight, synchronized multichannel audio and flight logs were recorded. Ground-truth labels consisted of leader bearing and horizontal range expressed in the follower body frame and derived from RTK measurements. Bearing convention used \SI{0}{\degree} for the leader lying along the follower UAV's forward direction, with angles wrapped to \SIrange{-180}{180}{\degree}. Bearing was interpolated at the center time of each acoustic window by linearly interpolating its sine and cosine components and then reconstructing the angle with \(\operatorname{atan2}\) to avoid discontinuities at the wrap boundary. Range was linearly interpolated at the same window-center time. Samples outside the \SIrange{1.0}{10.0}{\meter} range interval were excluded.

The final dataset includes \SI{140}{\minute} of flight data. Data was split by flight run to prevent temporal leakage between training and evaluation. Held-out flight runs were reserved exclusively for final testing, whereas the remaining runs were divided into training and validation subsets for model training and hyperparameter selection. During training, this pool was randomly divided into \SI{75}{\percent} training samples and \SI{25}{\percent} validation samples with a fixed random seed for checkpoint selection. The training split was then rebalanced by bearing bin through resampling to reduce over-representation of common relative headings.

\subsection*{Neural network architecture and learning framework}

SonicFly estimates leader bearing and range using a convolutional neural network designed for aeroacoustic relative-state regression. Bearing and range were selected as output targets because they form a compact follower-centered representation that can be directly used for pursuit control while remaining independent of global reference frames. Directional acoustic cues across the microphone array primarily provide bearing-related information, whereas distance-dependent attenuation, spectral structure, and inter-channel level differences provide range-related information. Unlike global Cartesian coordinates, this representation does not require a shared world frame or inter-vehicle communication, and it can be directly validated against RTK-derived relative pose.

Each acoustic window was converted into a normalized time-frequency representation, where the channel dimension contains stacked acoustic features from the microphone array, the frequency dimension corresponds to short-time spectral bins, and the temporal dimension spans consecutive acoustic frames. For window \(k\), the label vector is
\begin{equation}
\vect{y}_k=[s_k,c_k,\tilde r_k]^\top,
\end{equation}
where \(s_k\) is \(\sin\theta_k\), \(c_k\) is \(\cos\theta_k\), \(\theta_k\) is the leader bearing in the follower's body frame, and \(\tilde r_k\) is the normalized logarithmic range. We used a sine-cosine bearing representation to avoid the angular discontinuity at \(\pm\SI{180}{\degree}\), and a log-range label to reduce scale imbalance between near and far separations.

The neural estimator consists of a two-dimensional convolutional residual backbone followed by separate bearing and range regression heads. The backbone downsamples primarily along the frequency axis while preserving temporal resolution, compressing redundant spectral detail while retaining short-time temporal structure such as harmonic modulation and amplitude variation. The final feature map is pooled and projected to a compact latent representation. The bearing head outputs two values that are normalized to unit length to form \((\sin\theta,\cos\theta)\), and the range head outputs normalized log-range.

The model was trained with a multi-objective loss defined on bearing, range, and the implied relative Cartesian position. Bearing error was computed using circular angular distance, whereas range was optimized in normalized log-range space and augmented with a metric range-error term after denormalization. Because long-range underestimation can cause the follower UAV to lag behind the leader during pursuit, the range loss upweighted distant samples and penalized long-range underestimation. A Cartesian position term further aligned training with the downstream pursuit objective. Training used AdamW optimization with validation-based early stopping. Full architecture parameters, loss weights, and training hyperparameters are provided in Supplementary Materials.

\subsection*{Uncertainty management for a robust tracker} \label{sec:robustTracker}

Embodied passive aeroacoustic perception is subject to substantial uncertainty arising from ego-acoustic interference, environmental disturbances, and rapidly changing source-receiver geometry. Consequently, model predictions from our neural estimator may occasionally become unreliable despite strong average performance and our broad data coverage. Moreover, since our neural estimator was trained to produce one bearing--range estimate for each acoustic window, it does not by itself enforce a temporally consistent state estimate or a calibrated uncertainty model for each prediction. We therefore treat the neural network's outputs as uncertain predictions rather than direct state estimates and introduce a confidence-aware tracking framework to manage uncertainty during deployment.

For each prediction, a confidence score is computed from bearing consistency, range plausibility, and temporal continuity. The score is designed to identify measurements that deviate substantially from expected geometric or temporal behavior. Confidence hysteresis is then applied to prevent rapid switching between accepted and rejected measurements. Specifically, the confidence score is not a calibrated probability from the network. Instead, it is a deployment-time heuristic computed from the normalized prediction \(\hat{\vect{y}}_k=[\hat{s}_k,\hat{c}_k,\hat{\tilde r}_k]^\top\), where \(\hat{s}_k\) and \(\hat{c}_k\) are the predicted sine and cosine bearing components and \(\hat{\tilde r}_k\) is the normalized log-range estimate. The bearing and range confidence terms are
\begin{equation}
\begin{aligned}
\alpha_{\theta,k}
&=
\exp\!\left(
-\frac{
\left(\sqrt{\hat{s}_k^2+\hat{c}_k^2}-1\right)^2
}{2\sigma_{\theta,m}^2}
\right)
\eta_{\theta,k}
\exp\!\left[-\left(\frac{\Delta\theta_k}{\sigma_{\theta,t}}\right)^2\right],\\
\alpha_{r,k}
&=
\exp\!\left[-\left(\frac{|\hat{\tilde r}_k|}{\sigma_{r,d}}\right)^2\right]
\eta_{r,k}
\exp\!\left[-\left(\frac{|\hat{\tilde r}_k-\hat{\tilde r}_{k-1}|}{\sigma_{r,t}}\right)^2\right],
\end{aligned}
\end{equation}
where \(\Delta\theta_k\) is the frame-to-frame circular bearing change. The \(\sigma\) terms set the tolerance to deviations from a unit bearing vector, frame-to-frame bearing changes, distributional range outliers, and frame-to-frame range changes. The attenuation terms \(\eta_{\theta,k}\) and \(\eta_{r,k}\) reduce confidence for near-saturated bearing outputs and extreme normalized log-range values, respectively. The temporal terms are set to one when no previous prediction is available. The overall score used for gating is
\begin{equation}
\alpha_k = \left(w_\theta \alpha_{\theta,k}+w_r \alpha_{r,k}\right)^\gamma.
\end{equation}
Here, \(w_\theta\) and \(w_r\) are the bearing and range fusion weights, \(\gamma\) is a sharpening exponent that slightly suppresses ambiguous mid-confidence scores, and \(\alpha_k\) denotes the scalar confidence score at time step \(k\). All onboard acoustic estimation and outdoor pursuit results use the same confidence-score formulation, with fixed deployment parameters selected using the validation split from the training-validation dataset and kept unchanged for the final test set and reported field experiments. The hysteresis state transition is
\begin{equation}
g_k =
\begin{cases}
1, & g_{k-1}=0 \ \mathrm{and}\ \alpha_k \geq T_{\mathrm{high}}, \\
1, & g_{k-1}=1 \ \mathrm{and}\ \alpha_k \geq T_{\mathrm{low}}, \\
0, & \mathrm{otherwise},
\end{cases}
\end{equation}
where \(g_k \in \{0,1\}\) represents the gate state. This mechanism ensures that measurements are fused only when they remain consistently reliable, thereby preventing destabilizing ON/OFF chattering behavior in the feedback signal. When a prediction does not pass the gate, it is not used to update the filter. Instead, the tracker keeps the most recent published estimate when available.

Accepted model predictions are then represented as bearing--range measurements \(\vect{z}=[\theta,r]^\top\) and fused with an extended Kalman filter (EKF) \cite{kalman1960new} using a local Cartesian constant-velocity state \(\vect{x}=[p_x,p_y,v_x,v_y]^\top\), where \(p_x\) and \(p_y\) denote the leader source position relative to the follower in the horizontal plane. The measurement model is
\begin{equation}
h(\vect{x})=
\begin{bmatrix}
\mathrm{atan2}(p_y,p_x)\\
\sqrt{p_x^2+p_y^2}
\end{bmatrix}.
\end{equation}
We compute the innovation
\begin{equation}
\gvect{\nu}=\vect{z}-h(\hat{\vect{x}}_{t|t-1})
\end{equation}
and evaluate the Mahalanobis innovation distance
\begin{equation}
d^2=\gvect{\nu}^\top \mat{S}^{-1}\gvect{\nu},
\end{equation}
where
\begin{equation}
\mat{S}=\mat{H}\mat{P}_{t|t-1}\mat{H}^\top+\mat{R}.
\end{equation}
is the innovation covariance, and  \(\mat{H}\), \(\mat{P}_{t|t-1}\), and \(\mat{R}\) denote the measurement Jacobian, predicted state covariance, and measurement noise covariance, respectively. Measurements with large innovation distance are automatically down-weighted during the EKF update step, reducing the influence of transient acoustic prediction outliers while preserving temporal continuity.

The resulting Kalman-filtered estimate is passed directly to the pursuit controller. This strategy combines confidence hysteresis with innovation-aware Kalman updates to suppress unreliable model predictions while preserving a temporally continuous relative-state estimate. We will describe the pursuit controller in the next section.

\subsection*{Aeroacoustic source pursuit controller} \label{sec:control_strategy}
For drone-to-drone pursuit, the follower must regulate its motion to minimize its relative distance to the leader sound source while maintaining collision-safe separation. We achieve this behavior using a spring--mass--damper-inspired controller, which provides an intuitive and physically interpretable framework for stable interaction. Specifically, the control input is formulated as a velocity command generated by a virtual impedance model, where the resulting virtual force is proportional to both the relative displacement and its rate of change.

We define a closed ball \(\mathcal{B}_{r_0}(\mathbf{p}_s)=\{\mathbf{q}\in\mathbb{R}^2\mid \|\mathbf{q}-\mathbf{p}_s\|\leq r_0\}\) centered at the acoustic source position \(\mathbf{p}_s\) with radius \(r_0\), and denote its boundary as \(\partial\mathcal{B}_{r_0}(\mathbf{p}_s)\). The control objective is to regulate the drone toward this boundary. When the drone is outside the ball (\(\|\mathbf{p}-\mathbf{p}_s\|>r_0\)), an attractive velocity is generated toward \(\partial\mathcal{B}_{r_0}(\mathbf{p}_s)\). Conversely, when it is inside (\(\|\mathbf{p}-\mathbf{p}_s\|<r_0\)), a repulsive velocity drives it outward toward the same boundary. The commanded velocity is thus defined as
\begin{equation}
\vect{v} = -K_p \left( \|\vect{p}-\vect{p}_s\| - r_0 \right)\frac{\vect{p}-\vect{p}_s}{\|\vect{p}-\vect{p}_s\|} - K_d \dot{\vect{p}},
\end{equation}
where \(K_p\) and \(K_d\) denote the proportional and derivative gains, respectively. The proportional term regulates attraction or repulsion relative to the desired separation distance \(r_0\), while the derivative term damps the motion according to the drone velocity \(\dot{\vect{p}}\), reducing oscillatory behavior and improving pursuit stability. Consequently, the overall controller induces smooth attraction toward the source at long range and repulsion at close proximity, ensuring both convergence and safety. Rather than employing a relative-velocity damping term, we use only the drone velocity in the derivative component to avoid introducing additional sensitivity to estimation errors and temporal fluctuations in the acoustic source state. This hybrid interaction model enables robust and stable pursuit in dynamic and uncertain aeroacoustic environments. Successful pursuit under this controller indicates that the aeroacoustic estimates produced by SonicFly are sufficiently informative to support real-time aerial pursuit behavior.

\bibliography{bibliography} % for a file named science_template.bib
\bibliographystyle{sciencemag}

\section*{Acknowledgments}
The authors would like to thank Tate Staples and Jiaxun Liu for early explorations and Ben Borger for helpful discussions.

\paragraph*{Funding:}
Research was sponsored by the U.S. Army Combat Capabilities Development Command Army Research Laboratory (ARL) and was accomplished under Contract No. W911QX23C0011. The views and conclusions contained in this document are those of the authors and should not be interpreted as representing the official policies, either expressed or implied, of ARL. The U.S. Government is authorized to reproduce and distribute reprints for Government purposes notwithstanding any copyright notation herein. This work was also supported by DARPA TIAMAT program under award HR00112490419 and ARO under award W911NF2410405.

\paragraph*{Author contributions:}
B.C., Y.L., R.P., L.L., and N.R. conceived and designed the research. Y.L., R.P., L.L., and N.R. designed and performed physical experiments. B.C., Y.L., R.P., and L.L. analyzed data and wrote the manuscript. All authors provided feedback to the manuscript.

\paragraph*{Competing interests:}
There are no competing interests to declare.

\paragraph*{Data and materials availability:} All datasets, open-source code, and pretrained weights are available at \url{https://github.com/generalroboticslab/SonicFly}.

% %%%%%%%%%%%%%%%% SUPPLEMENT LIST %%%%%%%%%%%%%%%

% % List the contents of your Supplementary Materials, including the numbers of any
% % supplementary figures, tables, external data files etc. and any references that are
% % cited only in the supplement. In this example, refs. 7-8 are cited only in the supplement.
% % Fill out your numbers accordingly and delete any lines that aren't applicable.
% \subsection*{Supplementary materials}
% Materials and Methods\\
% Supplementary Text\\
% Figs. S1 to S3\\
% Tables S1 to S4\\
% References \textit{(7-\arabic{enumiv})}\\ % automatically fills out the last reference number
% % (filling out the other numbers automatically is possible but fiddly and liable to break)
% Movie S1\\
% Data S1

% %%%%%%%%%%%%%%%% END OF MAIN TEXT %%%%%%%%%%%%%%%

\newpage

%%%%%%%%%%%%%%%% START OF SUPPLEMENT %%%%%%%%%%%%%%%

% Figures, tables, equations and pages in the supplement are numbered S1, S2 etc.
\renewcommand{\thefigure}{S\arabic{figure}}
\renewcommand{\thetable}{S\arabic{table}}
\renewcommand{\theequation}{S\arabic{equation}}
\renewcommand{\thepage}{S\arabic{page}}
\setcounter{figure}{0}
\setcounter{table}{0}
\setcounter{equation}{0}
\setcounter{page}{1} % not 0 as \newpage already started a supplementary page
% References continue the numbering from the main text.

%%%%%%%%%%%%%%%% SUPPLEMENT TITLE PAGE %%%%%%%%%%%%%%%

\begin{center}
\section*{Supplementary Materials for\\ \scititle}

% Author list for the supplement
% Indicate the corresponding authors, but do NOT include institutions here
% It would be nice if the template auto-generated this, but doing so is complicated...
Yanbaihui Liu,
Ravi Prakash,
Li-Yu Lo,
Nils Roede,
Boyuan Chen$^{\ast}$\\
\small$^\ast$To whom correspondence should be addressed; E-mail: boyuan.chen@duke.edu.
\end{center}

% Fill out the numbers for each type of supplementary material,
% and delete any lines that aren't applicable.
% These are just example numbers that don't match the rest of this template.
\subsubsection*{This PDF file includes:}
Materials and Methods\\
Tables S1 to S2

% \subsubsection*{Other Supplementary Materials for this manuscript:}
% Movies S1 to S2\\
% Data S1 to S2

\newpage

%%%%%%%%%%%%%%%% MATERIALS AND METHODS %%%%%%%%%%%%%%%

\subsection*{Materials and Methods}

% The Materials and Methods section should contain details of the samples measured,
% experiments performed, observations taken, simulations run, data analysis, statistical methods etc.
% Give enough detail for any competent researcher in your field to fully reproduce the results.

% To refer to this section from the main text, use the numbered note in the reference list \cite{methods}.
% Refer to figures and tables in the same way as in the main text but now all capitalized e.g.
% Fig.~\ref{fig:example}, Table~\ref{tab:example},
% Fig.~\ref{fig:sup_example} and Table~\ref{tab:sup_example}.
% Cite references in the usual way \cite{example2},
% including any that are only cited in the supplement \cite{sm_example,conference_example}.

% The numbering of figures, tables, equations and pages has been reset to start from S1, as in
% \begin{equation}
% 	\cos(2\theta) = \cos^2\theta - \sin^2\theta.
% 	\label{eq:sup_example} % Use a logical label
% \end{equation}

% \subsubsection*{Example supplement heading}

% The two main sections of the supplement can be split up using headings.

%%%%%%%%%%%%%%%% SUPPLEMENTARY TEXT %%%%%%%%%%%%%%%
\subsubsection*{Comparison of microphones} \label{SI:mic_comparison}

We assumed a nominal sound pressure level (SPL) of \SI{80}{dB} generated by the leader drone, estimated in real-world conditions heuristically. Microphones were evaluated to possess sufficient SNR compared to ambient environment noise at approximately \SI{40}{\meter} distance.

\begin{table}[htbp]
    \centering
    \caption{Comparison of candidate MEMS microphones considered for acoustic sensing.}
    \label{tab:sensor_selection}
    \renewcommand{\arraystretch}{1.15}
    \small
    \begin{tabular}{p{2.7cm} p{1.7cm} p{2.0cm} p{1.5cm} p{3.0cm} p{2.0cm}}
        \toprule
        \textbf{Sensor} & 
        \textbf{Type} & 
        \textbf{Sensitivity} & 
        \textbf{SNR} & 
        \textbf{Frequency range} & 
        \textbf{Estimated distance} \\
        & & \textbf{(\si{dBFS})} & \textbf{(\si{dBA})} & & \textbf{(\si{\meter})} \\
        \midrule
        ReSpeaker Mic v2.0 & Digital & $-26$ & 64 & \SI{100}{\hertz}--\SI{10}{\kilo\hertz} & 96.3 \\
        INMP441 & Digital, omnidirectional & $-26$ & 61 & \SI{60}{\hertz}--\SI{15}{\kilo\hertz} & 68.3 \\
        \textbf{ICS-43434} & \textbf{Digital} & $\mathbf{-26}$ & \textbf{65} & \textbf{\SI{23}{\hertz}--\SI{51.6}{\kilo\hertz}} & \textbf{107.9} \\
        \bottomrule
    \end{tabular}
\end{table}

\newpage

\subsubsection*{Acoustic feature construction and detailed architecture}

Audio was segmented into \SI{0.25}{\second} windows with a \SI{0.12}{\second} hop. For the \SI{12}{\kilo\hertz} audio stream, this corresponds to 3000 audio samples per window and 1440 samples between adjacent windows. For each window and each microphone channel, we computed a centered short-time Fourier transform (STFT) with \(N_{\mathrm{FFT}}=512\), an STFT hop length of 128 samples, and centered padding. This produced 257 frequency bins and 24 temporal frames per acoustic window.

Each input tensor contained 13 acoustic feature channels, which consisted of 4 log-magnitude spectrograms, 3 interaural phase difference (IPD) sine channels, 3 IPD cosine channels, and 3 interaural level difference (ILD) channels, all computed relative to microphone 0. Log-magnitude spectrogram features were retained in the selected \SIrange{1}{2}{\kilo\hertz} band, whereas IPD and ILD features were retained from \SI{0}{\hertz} to \SI{2300}{\hertz}. Channels outside the corresponding bands were masked before normalization.

The \SIrange{1}{2}{\kilo\hertz} log-magnitude spectrogram range was selected to emphasize frequency regions in which the leader drone remains distinguishable from follower ego-acoustics while suppressing very-low-frequency platform vibration and less stable high-frequency components. For IPD and ILD, we retained \SIrange{0}{2300}{\hertz} to preserve inter-microphone phase and level cues over the frequency region where the leader-to-follower contrast remains informative. This upper bound is also consistent with the adjacent-pair half-wavelength spatial-aliasing limit of approximately \SI{2.3}{\kilo\hertz} for the \SI{75}{\milli\meter} square microphone layout.

The resulting input had shape
\begin{equation}
\mat{X} \in \mathbb{R}^{13 \times 257 \times 24}.
\end{equation}
Audio features were normalized using channel-frequency cepstral mean and variance normalization fit on the training runs. The leader range was represented as log-range and standardized using training-set statistics. For sample \(i\), bearing was represented as \(s_i=\sin\theta_i\) and \(c_i=\cos\theta_i\), yielding the label vector
\begin{equation}
\vect{y}_i=[s_i,c_i,\tilde r_i]^\top ,
\end{equation}
where \(\tilde r_i\) denotes normalized log-range.

The detailed architecture of SonicNet is summarized in Table~\ref{tab:sonicnet_architecture}. The bearing and range heads each used a two-layer multilayer perceptron with a 128-dimensional hidden layer, Gaussian Error Linear Unit (GELU) activation, and dropout. The bearing output was normalized to unit length:
\begin{equation}
(\hat s,\hat c) =
\frac{(z_s,z_c)}{\sqrt{z_s^2+z_c^2+\epsilon}},
\end{equation}
where \((z_s,z_c)\) are the raw bearing-head outputs. The network contained 1.73 million trainable parameters in total.

\begin{table}[h]
\centering
\caption{\textbf{Neural network architecture.} Input dimensions are channel \(\times\) frequency \(\times\) time.}
\label{tab:sonicnet_architecture}
\begin{tabular}{llll}
\hline
Module & Operation & Channels & Output size \\
\hline
Input & Acoustic features & 13 & \(13 \times 257 \times 24\) \\
Stem & \(5 \times 5\) conv, stride \((1,1)\) & 64 & \(64 \times 257 \times 24\) \\
Residual stage 1 & residual conv, stride \((2,1)\) & 96 & \(96 \times 129 \times 24\) \\
Residual stage 2 & residual conv, stride \((2,1)\) & 128 & \(128 \times 65 \times 24\) \\
Residual stage 3 & residual conv, stride \((2,1)\) & 160 & \(160 \times 33 \times 24\) \\
Residual stage 4 & residual conv, stride \((2,1)\) & 192 & \(192 \times 17 \times 24\) \\
Projection & frequency pooling, temporal conv, global pooling & 256 & \(256\) \\
Bearing head & MLP + unit normalization & 2 & \((\sin\theta,\cos\theta)\) \\
Range head & MLP & 1 & \(\tilde r\) \\
\hline
\end{tabular}
\end{table}

\clearpage

\subsubsection*{Training loss}

The model was trained with a supervised bearing--range--position loss. Bearing error was computed as a circular angular residual from the predicted and ground-truth sine-cosine vectors. For sample \(i\), let \((\hat{s}_i,\hat{c}_i)\) be the predicted bearing vector and \((s_i,c_i)\) the ground truth bearing vector. The predicted bearing vector was normalized to unit length before evaluating the loss. The angular residual was
\begin{equation}
\Delta\theta_i =
\operatorname{atan2}
(\hat{s}_i c_i-\hat{c}_i s_i,\,
 \hat{c}_i c_i+\hat{s}_i s_i).
\end{equation}
The bearing loss combined the wrapped angular error with a range-aware chord term,
\begin{equation}
\mathcal{L}_{\theta}
=
\frac{1}{B}\sum_i |\Delta\theta_i|
+
0.10
\frac{1}{B}\sum_i
r_i^{c}\,2\sin\left(\frac{|\Delta\theta_i|}{2}\right),
\end{equation}
where \(B\) denotes the mini-batch size. The clipped metric range \(r_i^{c}\) is defined below. This auxiliary term increases the cost of bearing errors for distant targets, where the same angular error corresponds to a larger Cartesian position error.

Range was optimized in normalized log-range space, with an additional metric-relative-error term after denormalization. Let \(\hat{\tilde r}_i\) and \(\tilde r_i\) denote predicted and ground-truth normalized log-range values. For the range weighting and relative-error term, the corresponding metric ranges \(\hat r_i^{c}\) and \(r_i^{c}\) were obtained by denormalizing the log-range values and clipping them to the physically plausible interval \SIrange{1}{12}{\meter}.

The range sample weight \(w_i\) was set to 1 by default. It was multiplied by 1.2 for short-range samples \((r_i^{c}<\SI{3.5}{\meter})\), multiplied by 2.2 for long-range samples \((r_i^{c}\geq\SI{7.0}{\meter})\), and further multiplied by 1.2 when the leader was at long range and underestimated by more than \SI{0.3}{\meter}:
\begin{equation}
r_i^{c} \geq \SI{7.0}{\meter}
\quad \mathrm{and} \quad
r_i^{c}-\hat r_i^{c} > \SI{0.3}{\meter}.
\end{equation}
% \el{self-note: maybe explain why different range multipliers?}
The short-range multiplier improves collision-relevant accuracy near the desired standoff distance, whereas the long-range and underestimation multipliers compensate for the pursuit risk introduced by attenuated acoustic observations at large separations.

The range loss was
\begin{equation}
\mathcal{L}_{r}
=
0.65 \frac{1}{B}\sum_i w_i|\hat{\tilde r}_i-\tilde r_i|
+
0.35 \frac{1}{B}\sum_i w_i
\frac{|\hat r_i^{c}-r_i^{c}|}{r_i^{c}+\epsilon}.
\end{equation}
To align training with the downstream pursuit objective, we also penalized error in the predicted relative Cartesian position. For this term, the predicted position range \(\hat r_i^{xy}\) was computed from the denormalized predicted log-range with a wider numerical clamp on log-range, while the ground-truth position used \(r_i^{c}\). Using \(x=r\cos\theta\) and \(y=r\sin\theta\), the position loss was
\begin{equation}
\mathcal{L}_{xy}
=
\frac{1}{B}\sum_i
\operatorname{SmoothL1}
\left(
\left[
\begin{array}{c}
\hat r_i^{xy} \hat c_i\\
\hat r_i^{xy} \hat s_i
\end{array}
\right]
-
\left[
\begin{array}{c}
r_i^{c} c_i\\
r_i^{c} s_i
\end{array}
\right]
\right),
\end{equation}
with the SmoothL1 loss summed over the two Cartesian coordinates. The total loss was
\begin{equation}
\mathcal{L}
=
w_{\theta}\mathcal{L}_{\theta}
+
w_r\mathcal{L}_r
+
0.35\mathcal{L}_{xy}.
\end{equation}
The task weights \(w_{\theta}\) and \(w_r\) were computed from inverse label variance on the training set, and the range task weight was further multiplied by 2.5 to emphasize range accuracy during pursuit.

\subsubsection*{Training protocol and model selection}

The training-validation pool was randomly divided into \SI{75}{\percent} training samples and \SI{25}{\percent} validation samples using a fixed random seed. Training used AdamW with batch size 64, initial learning rate \(5\times10^{-3}\), weight decay \(10^{-3}\), gradient clipping at 1.0, cosine learning-rate decay, and 5 warmup epochs. Training was run for at most 270 epochs with early stopping patience of 15 epochs. Mixed-precision training was used when CUDA was available. The checkpoint used for evaluation was selected from the validation set using a combined criterion over validation loss, bearing error, range mean absolute error, and Cartesian position mean absolute error, with weights 0.10, 0.35, 0.20, and 0.35, respectively.

Confidence gating, Kalman filtering, and median filtering were not part of network training. The network was trained as a single-window supervised bearing-range regressor. The filtering stages were applied only after the neural network inference in the downstream acoustic tracker.

\clearpage % Clear all remaining figures and tables then start a new page

% \paragraph{Caption for Movie S1.}
% \textbf{All captions must start with a short bold sentence, acting as a title.}
% Then explain what is shown in the supplementary video file.
% Give as much detail as you would for a figure e.g. explain axes, color maps etc.
% If the video is an animated equivalent of one of the static figures, state e.g.
% `Animated version of Figure~\ref{fig:example}.'

% \paragraph{Caption for Data S1.}
% \textbf{All captions must start with a short bold sentence, acting as a title.}
% Then explain what is included in the supplementary data file.
% Give as much detail as you would for a table e.g. explain the meaning of every column,
% units used, any special notation etc.

%%%%%%%%%%%%%%%% SUPPLEMENTARY REFERENCES %%%%%%%%%%%%%%%

% Do NOT include a reference list in the supplement.
% All references must be in a single list at the end of the main text.
% The copyeditors will ensure that the correct reference list appears with each version of the paper
% (print, HTML, PDF, mobile app, metadata for bibliographic databases etc.)

\end{document}